\documentclass[10pt,twocolumn,letterpaper,table,hyphens]{article}

\usepackage{color,xcolor}
\usepackage{epsfig}
\usepackage{graphicx}

\usepackage{comment}

\usepackage{pifont}

\usepackage{microtype}
\usepackage[font=small]{caption}
\usepackage{arydshln}
\usepackage{tabularx}
\usepackage{adjustbox}
\usepackage{array}
\usepackage{booktabs}
\usepackage{colortbl}
\usepackage{float,wrapfig}
\usepackage{hhline}
\usepackage{multirow}
\usepackage{subcaption} %
\usepackage[percent]{overpic}

\usepackage{stmaryrd}

\usepackage{amsmath,amsfonts,amssymb}
\usepackage{bm}
\usepackage{nicefrac}
\usepackage{microtype}
\usepackage{dsfont}
\usepackage{changepage}
\usepackage{extramarks}
\usepackage{fancyhdr}
\usepackage{setspace}
\usepackage{soul}
\usepackage{xspace}

\usepackage[pagebackref=true,breaklinks=true,letterpaper=true,colorlinks,bookmarks=false,hypertexnames=false]{hyperref}
\usepackage{url}

\usepackage{algorithm, algorithmic}
\usepackage{enumitem}
\usepackage[title]{appendix}

\usepackage{amsmath,amsfonts,bm,relsize,dsfont}
\definecolor{MyDarkBlue}{rgb}{0,0.08,1}
\definecolor{MyDarkGreen}{rgb}{0.02,0.6,0.02}
\definecolor{MyDarkRed}{rgb}{0.8,0.02,0.02}
\definecolor{MyDarkOrange}{rgb}{0.40,0.2,0.02}
\definecolor{MyPurple}{RGB}{111,0,255}
\definecolor{MyRed}{rgb}{1.0,0.0,0.0}
\definecolor{MyGold}{rgb}{0.75,0.6,0.12}
\definecolor{MyDarkgray}{rgb}{0.66, 0.66, 0.66}
\definecolor{MyDarkCyan}{rgb}{0.05, 0.55, 0.45}
\definecolor{MyBlack}{rgb}{0., 0., 0.}
\definecolor{MyMagenta}{rgb}{1., 0., 1.}
\definecolor{BerkeleyYellow}{RGB}{255,204,41}
\definecolor{BerkeleyLightBlue}{RGB}{94,146,221}
\definecolor{BkDarkBlue}{rgb}{.05,.07,.353}

\definecolor{MyDarkGray2}{rgb}{0.6, 0.6, 0.6}
\newcommand{\gray}[1]{\textcolor{MyDarkGray2}{#1}}

\newcommand{\camready}[1]{#1}
\newcommand{\camrdy}[1]{#1}

\newcommand{\suppmat}[1]{\textcolor{MyBlack}{#1}}

\newcommand{\fid}{Fr\'echet Inception Distance\xspace}

\def\rvx{{\mathbf{x}}}
\def\rvy{{\mathbf{y}}}
\def\rvz{{\mathbf{z}}}

\newcommand{\reffig}[1]{Figure~\ref{fig:#1}}
\newcommand{\refsec}[1]{Section~\ref{sec:#1}}

\newcommand{\reftbl}[1]{Table~\ref{tbl:#1}}

\newcommand{\refeq}[1]{Eqn.~\ref{eq:#1}}

\newcommand{\lblfig}[1]{\label{fig:#1}}
\newcommand{\lblsec}[1]{\label{sec:#1}}
\newcommand{\lbleq}[1]{\label{eq:#1}}

\newcommand{\ignorethis}[1]{}

\newcommand{\myparagraph}[1]{\vspace{1pt} \noindent \textbf{#1} \ }

\def\1{\bm{1}}

\newcommand{\image}{{\rvx}}
\newcommand{\latent}{{\rvz}}
\newcommand{\images}{{\mathcal{X}}}
\newcommand{\imagedist}{{p_{\text{data}}(\image)}}
\newcommand{\latentdist}{{p(\latent)}}
\newcommand{\imageD}{{D_X}}
\newcommand{\Fnet}{{F}}
\newcommand{\sketch}{{\rvy}}
\newcommand{\sketches}{{\mathcal{Y}}}
\newcommand{\sketchdist}{{p_{\text{data}}(\sketch)}}
\newcommand{\sketchD}{{D_Y}}
\newcommand{\modelold}{{G(\rvz; \theta)}}
\newcommand{\modelnew}{{G(\rvz; \theta')}}
\newcommand{\losssketch}{{\mathcal{L}_{\text{sketch}}}}
\newcommand{\lossimage}{{\mathcal{L}_{\text{image}}}}
\newcommand{\lossweight}{{\mathcal{L}_{\text{weight}}}}

\newcommand{\method}{{GAN Sketching}}

\newcolumntype{L}[1]{>{\raggedright\let\newline\\\arraybackslash\hspace{0pt}}m{#1}}
\newcolumntype{C}[1]{>{\centering\let\newline\\\arraybackslash\hspace{0pt}}m{#1}}
\newcolumntype{R}[1]{>{\raggedleft\let\newline\\\arraybackslash\hspace{0pt}}m{#1}}

\newcommand{\ignore}[1]{}

\renewcommand*{\thefootnote}{\arabic{footnote}}

\makeatletter
\DeclareRobustCommand\onedot{\futurelet\@let@token\@onedot}
\def\@onedot{\ifx\@let@token.\else.\null\fi\xspace}

\def\etal{\emph{et al}\onedot}
\makeatother

\usepackage{iccv}
\usepackage{times}
\usepackage{epsfig}
\usepackage{graphicx}
\usepackage{amsmath}
\usepackage{amssymb}

\usepackage[pagebackref=true,breaklinks=true,letterpaper=true,colorlinks,bookmarks=false]{hyperref}

\iccvfinalcopy %

\def\iccvPaperID{1622} %

\ificcvfinal\fi

\begin{document}

\title{Sketch Your Own GAN}

\author{Sheng-Yu Wang\textsuperscript{1}
\qquad
David Bau\textsuperscript{2}
\qquad
Jun-Yan Zhu\textsuperscript{1}
\\
\textsuperscript{1}Carnegie Mellon University
\qquad
\textsuperscript{2}MIT CSAIL
}

\twocolumn[{%
\renewcommand\twocolumn[1][]{#1}%
\maketitle

\begin{center}
    \centering
    \includegraphics[width=0.95\linewidth]{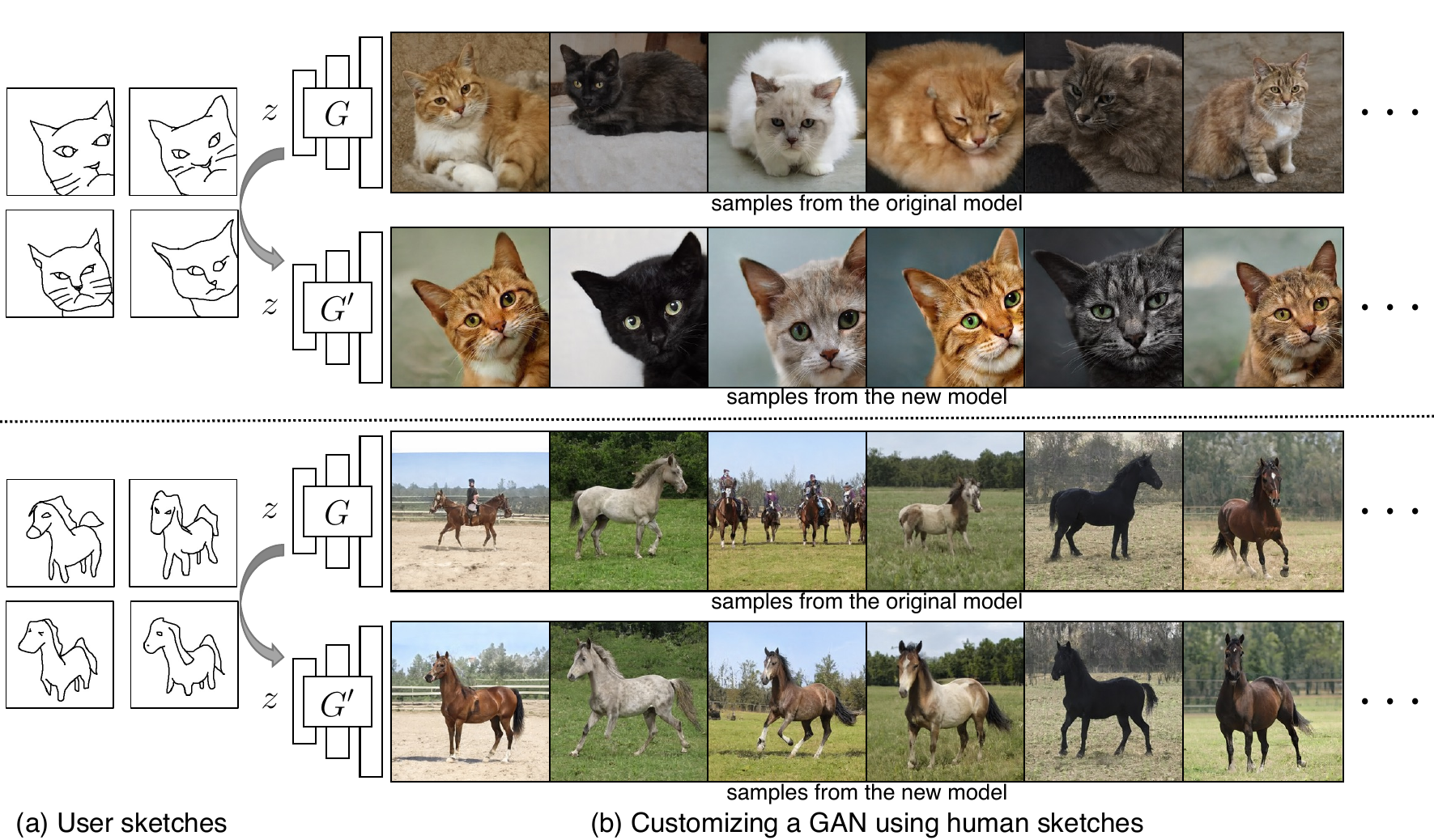}
    \captionof{figure}{\textbf{Customizing a GAN with one or more human sketches.} Our method takes in one or a few hand-drawn sketches (a) and modifies an off-the-shelf GAN to match the input sketch (b). The same noise $z$ is used for both the original model $G$ and modified model $G'$. While our new model changes an object's shape and pose, other visual cues, such as color, texture, and background, are faithfully preserved after the modification. }%
    \label{fig:teaser}
\end{center}
\vspace{-5pt}
}]
 \maketitle

\begin{abstract}
\vspace{-10pt}
 
   Can a user create a deep generative model by sketching a single example? Traditionally, creating a GAN model has required the collection of a large-scale dataset of exemplars and specialized knowledge in deep learning. In contrast, sketching is possibly the most universally accessible way to convey a visual concept. 
  In this work, we present a method, \camready{\method}, for rewriting GANs with one or more sketches, to make GANs training easier for novice users. In particular, we change the weights of an original GAN model according to user sketches. We encourage the model's output to match the user sketches through a cross-domain adversarial loss. Furthermore, \camready{we explore different regularization methods to preserve the original model's diversity and image quality.}  Experiments have shown that our method can mold GANs to match shapes and poses specified by sketches while maintaining realism and diversity. Finally, we demonstrate a few applications of the resulting GAN, including latent space interpolation and image editing.

\end{abstract}

\vspace{-15pt}
\section{Introduction}
The power and promise of deep generative models such as GANs~\cite{goodfellow2014generative} lie in their ability to synthesize endless realistic, diverse, and novel content with minimal user effort.  The potential utility of these models continues to grow thanks to the increased quality and resolution of large-scale generative models~\cite{karras2020analyzing,brock2019large,razavi2019generating,ramesh2021zero} in recent years.

Nonetheless, the training of high-quality generative models demands high-performance computing platforms, 
putting the process out of reach for most users.  
Furthermore, training a high-quality model requires expensive large-scale data collection and careful pre-processing.  Commonly used datasets such as ImageNet~\cite{deng2009imagenet} and LSUN~\cite{yu2015lsun} require human annotation and manual filtering. The specialized FFHQ Face dataset~\cite{karras2019style} requires delicate face alignment and super-resolution pre-processing. Moreover, the technical effort is not trivial: developing an advanced generative model requires the domain knowledge~\cite{salimans2016improved,karras2020analyzing} of a team of experts, who often invest months or years into a single model on specific datasets.   

This leads to the question: how can an ordinary user create their own generative model?  A user creating artwork with cats might not want a generic model of cats, but a bespoke model of special cats in a particular desired pose: nearby, reclining, or all looking left.  To obtain such a customized model, must the user curate thousands of reclining left-looking cat images and then find an expert to invest months of time in model training and parameter tuning?

In this paper, we propose the task of creating a generative model from just a handful of hand-drawn sketches.  Ever since Ivan Sutherland's SketchPad~\cite{sutherland1964sketchpad}, computer scientists have recognized the usefulness of guiding computer-generated content using sketching interface.  This tradition has continued in the area of sketch-based image synthesis and 3D modeling~\cite{igarashi1999teddy,chen2009sketch2photo,isola2017image}.  But rather than creating a single image or a 3D shape from a sketch, we wish to understand if it is possible to create a generative model of realistic images from hand-drawn sketches. Unlike sketch-based content creation, where both the input and output are 2D or 3D visual data, in our case, the input is a 2D sketch and the output is a network with millions of opaque parameters that control algorithm behavior to make images. We ask: with such a different output domain, which parameters shall we update, and how? How do we know whether the model's output will resemble the user sketch?

In this paper, we aim to answer the above questions by developing a method to tailor a generative model to a small number of sketch exemplars provided by the user. To achieve this, we take advantage of off-the-shelf generative models pre-trained on large-scale data, and devise an approach to adjust a subset of the model weights to match the user sketches. We present a new cross-domain model fine-tuning method that encourages the new model to create images that resemble a user sketch, while preserving the color, texture, and background context of the original model. As shown in \reffig{teaser}, our method can change the object pose and zoom in cat faces with only four hand-drawn sketches.

We use our method to create several new customized GAN models, and we show that these modified models can be used for several applications such as generating new samples, interpolating between two generated images, as well as editing a natural photograph. Our method requires minimal user input. Instead of collecting a new dataset through manual filtering and image alignment, a user only needs to provide one or a few exemplar sketches for our method to work effectively. Finally, we benchmark our method to fully characterize its performance. \camready{%
\href{https://github.com/PeterWang512/GANSketching}{Code} and models are also available on our \href{https://peterwang512.github.io/GANSketching}{webpage}.}%

\vspace{-5pt}
\section{Related Works}
\myparagraph{Sketch based image retrieval and synthesis.}  
Retrieving images that resemble a human sketch has been extensively studied, including classic methods~\cite{eitz2010sketch,cao2011edgel,shrivastava2011data,lin20133d} that rely on feature descriptors, as well as more recent deep learning methods~\cite{yu2016sketch,sangkloy2016sketchy,liu2017deep,radenovic2018deep,ribeiro2020sketchformer}. The above sketch-based image retrieval (SBIR) techniques have powered sketch-based 3D modeling systems (e.g., Teddy~\cite{igarashi1999teddy}) as well as image synthesis systems, including Sketch2Photo~\cite{chen2009sketch2photo} and PhotoSketcher~\cite{eitz2011photosketcher}. These seminal works have further inspired deep learning solutions based on image-to-image translation~\cite{isola2017image,zhu2017unpaired,wang2018pix2pixHD} such as Scribbler~\cite{sangkloy2017scribbler}, SketchyGAN~\cite{chen2018sketchygan}, SketchyCOCO~\cite{gao2020sketchycoco}, and sketch-based face and hair editing~\cite{portenier2018faceshop,olszewski2020intuitive}. Other relevant work includes sketch recognition~\cite{eitz2012humans,yu2017sketch} and sketch generation~\cite{ha2018neural}. Collectively, the above methods have enabled a novice user to synthesize a \emph{single} natural photograph. %
In this work, we would like to employ the same intuitive interface for rewriting a generative model. Once done, the resulting model can produce an infinite number of new samples that resemble the input sketch. The models can be further used for random sampling, latent space interpolation, as well as natural photo editing. %

\myparagraph{Generative models for content creation.} After years of development, deep generative models are able to produce high-quality, high-resolution images~\cite{goodfellow2014generative,karras2019style,brock2019large,kingma2018glow}, powering a wide range of computer vision and graphics applications. Recent examples include image projection and editing with GANs~\cite{zhu2016generative,bau2019semantic,abdal2019image2stylegan,zhu2020domain,richardson2020encoding,patashnik2021styleclip}, image-to-image translation~\cite{isola2017image,liu2019few,park2019SPADE,lee2018diverse}, simulation-to-real~\cite{rao2020rl,shrivastava2017learning}, and domain adaptation~\cite{ganin2016domain,tzeng2017adversarial,hoffman2018cycada}. The advance of generative models comes at the cost of intensive computation~\cite{brock2019large,karras2020analyzing}, the construction of large-scale high-quality datasets~\cite{deng2009imagenet,yu2015lsun,karras2019style}, and domain expertise on model training~\cite{salimans2016improved,karras2020analyzing}. As a result, recent advanced models are often developed in research labs with abundant computing and human resources. Different from prior works, we aim to help novice users quickly customize their own models without tedious data collection and domain knowledge. We achieve it using a sketching interface and our cross-domain fine-tuning method. Similar to our work, model rewriting~\cite{bau2020rewriting} aims to change the rules of a pre-trained model through user interaction. Compared to the object copy-and-paste tool used in model rewriting, our sketching interface provides new capabilities. For example, it is much easier to describe the object shape, pose, and scene layout through quick sketching, compared to finding parts from different images and compositing them together. %

\myparagraph{Model fine-tuning.}%
To train a GAN model of a new dataset, researchers have fine-tuned the weights of a pre-trained generator and discriminator pair using transfer learning~\cite{donahue2014decaf,zeiler2014visualizing}. The fine-tuning can improve upon training from scratch~\cite{wang2018transferring}, but it can also easily overfit on the new training data. To avoid overfitting, %
several groups propose to limit the changes in model weights: Batch Statistic Adaptation preserves all weights except batch statistics~\cite{noguchi2019image}; Freeze-D freezes layers of the discriminator~\cite{mo2020freeze}; AdaFM preserves generator layers~\cite{zhao2020leveraging}; MineGAN keeps weights while altering latent-sampling~\cite{wang_2020_CVPR}; and Elastic Weight Consolidation~\cite{kirkpatrick2017overcoming} preserves weights based on Fisher information~\cite{li2020fewshot}.  The other technique is data augmentation, which has been proven effective for small-scale datasets~\cite{zhao2020diffaugment,karras2020ADA,tran2020towards,zhao2020image}. Different from the above works, our goal is \emph{not} to fine-tune weights to learn a model of sketches.  Instead, we aim to enable a user to create a new model of realistic images that builds upon the color, texture, and details of a pre-trained GAN, guided by the user-specified object shapes and poses from sketches.

\vspace{-5pt}
\section{Methods}
\lblsec{methods}

Two constraints make the creation of a GAN model from user sketches challenging.
First, as our goal is to simplify the user creation of a generative model, we must utilize only a very small amount of user-provided sketch data.  It would be unreasonable to require the user to supply hundreds or thousands of sketches; instead, we aim to be able to create a model using as few as one single sketch.

Second, as we aim to synthesize realistic images without requiring the user to create realistic images, the user-provided sketches are not drawn from the target domain.  This mismatch between training data (i.e., sketch) and model output (i.e., image) makes our problem setting drastically different from the traditional GAN training objective, which is to match the training data directly. In our setting, the goal is to create a model of realistic photographs where the shape and pose are guided by sketches --- but where the outputs are realistic images, rather than sketches.

To overcome the above challenges, we introduce a cross-domain adversarial loss using a domain-translation network (\refsec{cross_domain}). Unfortunately, simply using this loss dramatically changes the model's behavior and produces unrealistic results. To preserve the content of the original dataset as well as its diversity, we further train the model while applying \camready{image-space regularization (\refsec{image_reg})}. Finally, to alleviate model overfitting, we limit the updates to specific layers and use data augmentation in \refsec{opt}.

\begin{figure}
    \centering
    \includegraphics[width=1.\linewidth]{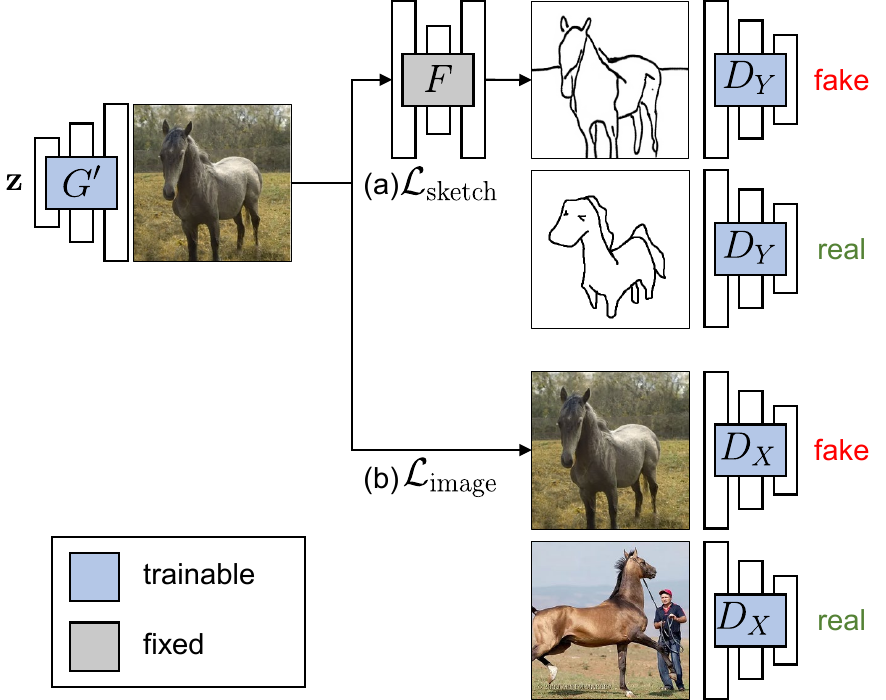}
    \vspace{-10pt}
    \caption{\textbf{Training procedure.} Our training consists of two major components. \textbf{(a) $\losssketch$:} the sketch discriminator $\sketchD$ classifies between fake and user sketches. A pre-trained mapping network $\Fnet$~\cite{li2019photo} is used to translate the output of our model $\modelnew$ to a fake sketch. \textbf{(b) $\lossimage$:} the image discriminator $\imageD$ classifies between fake and real images. Real images are sampled from the training set of the original model $\modelold$.}
    \lblfig{training}
    \vspace{-10pt}
\end{figure}

 \subsection{Cross-Domain Adversarial Learning}
\lblsec{cross_domain}

Let $\images, \sketches$ be the domains that consist of images and sketches, respectively. We have collected a large-scale set of training images $\image \sim \imagedist $ and a few human sketches $\sketch \sim \sketchdist$. We denote $\modelold$ as a pre-trained GAN that produces an image $\image$ from a low-dimensional code $\latent$, %
We wish to create a new GAN model $\modelnew$, whose output images still follow the same data distribution of $\images$, while the sketch version of the output images are similar to the data distribution of $\sketches$. 

Previous few-shot GAN algorithms fail to work in this setting as no ground truth images from new dataset are provided by the user. To address the challenge, we utilize a cross-domain image translation network from images to sketches $\Fnet:\images \rightarrow \sketches$. This network can be trained using input-output pairs such as photos and their sketch version. Alternatively, it can be learned through unpaired learning~\cite{zhu2017unpaired,huang2018multimodal}. Once the mapping network is pre-trained, we do not need the sketch-image ground truth pairs during our model creation. %

To bridge the gap between sketch training data and the image generative model, we introduce a cross-domain adversarial loss~\cite{goodfellow2014generative} to encourage the generated images to match the sketches $\mathcal{Y}$. Before passing into the discriminator, the output of the generator is transferred into a sketch by the pre-trained image-to-sketch network $\Fnet$.
\begin{equation}
\begin{aligned}
    \losssketch & = \mathbb{E}_{\sketch \sim \sketchdist} \log (\sketchD(\sketch)) \\
    & + \mathbb{E}_{\latent \sim \latentdist} \log (1-\sketchD(\Fnet(G(\latent)))),
\lbleq{sketch}
\end{aligned}
\end{equation}
Where we use Photosketch~\cite{li2019photo}  as our image to sketch network $\Fnet$.
Note that our method is able to generalize to sketch examples that are not from the original PhotoSketch training or test set (\refsec{quantitative_eval}). Despite the difference in sketching styles, the network still helps capture the overall shape of the object.

\subsection{Image Space Regularization}
\lblsec{image_reg}

We observe that using the loss on sketches alone leads to a drastic degradation in image quality and diversity in generation, as this loss only enforces the shape of the generated images to match the sketch.

To resolve this, we add a second adversarial loss that compares outputs to the training set of the original model.

\begin{equation}
\begin{aligned}
    \lossimage & = \mathbb{E}_{\image \sim \imagedist} \log (\imageD(\image)) \\
    & + \mathbb{E}_{\latent \sim \latentdist} \log (1-\imageD(G(\latent))).
\lbleq{image}
\end{aligned}
\end{equation}

A separate discriminator $\imageD$ is used  \camready{to preserve the image quality and diversity of model outputs while matching the user sketch.} %

\paragraph{Weight regularization as an alternative.}
\camready{We also experiment with weight regularization, where we explicitly penalize large changes in the weights using the following loss:} 

\begin{equation}
    \lossweight = ||\theta' - \theta||_1.
    \lbleq{weight}
\end{equation}

\camready{Although this regularization does not require the training set of the original model, we observe that applying weight regularization leads to slightly worse performance than image space regularization (\refeq{image}). See more detailed comparisons in \refsec{quantitative_eval}.}

\camready{A recent approach, Elastic Weight Consolidation (EWC)~\cite{kirkpatrick2017overcoming},  aims to overcome catastrophic forgetting and can be used to regularize few-shot GANs training~\cite{li2020fewshot}.  In our setting, we find that a simple L1-based weight regularization loss (\refeq{weight}) performs on par with EWC.  We report results of L1-based weight regularization for simplicity.}

\camready{We experiment with models trained with both image and weight regularization methods combined, and find that they do not outperform models trained with only image regularization. However, we note that applying either weight or image regularization is critical for balancing image quality and shape matching.} %

\subsection{Optimization}
\lblsec{opt}
Our full objective is:  
\begin{equation}
    \mathcal{L}= \mathcal{L}_{\text{sketch}} + \lambda_{\text{image}}\lossimage,
\end{equation}
\camready{with $\lambda_{\text{image}}=0.7$, controlling the importance of the image regularization term}.  %
We would like to learn a new set of weights $\modelnew$ with the following minimax objective: 
\begin{equation}
\theta'= \arg\min_{\theta'}\max_{\imageD, \sketchD} \mathcal{L}.
\end{equation}

\myparagraph{Which layers to edit.} To prevent model overfitting and accelerate fine-tuning speed, we only modify the weights of the mapping network of StyleGAN2~\cite{karras2020analyzing}, which essentially remaps $z \sim \mathcal{N}(0, I)$ to a different intermediate latent space ($\mathcal{W}$ space). We observe this to be effective, since modifying the mapping network is sufficient to obtain our target distribution, which is a subset of the original distribution. This choice has also been shown effective in previous few-shot GAN works~\cite{wang2019MineGAN}. \camready{We have also experimented with optimizing the entire generator, and observe that generated output contains severe artifacts.} %

\myparagraph{Pre-trained weights.} We use a pre-trained Photosketch network $\Fnet$ and we  fix the weights of $\Fnet$ throughout the training.  We optimize all the parameters for both $\imageD$, $\sketchD$ during training. $\imageD$, $\sketchD$ are initialized with the pre-trained weights from the original GAN. 
\suppmat{More training details are provided in Appendix~\ref{sec:impl_details}.}

\myparagraph{Data augmentation.}
\camready{We experiment with differentiable augmentation~\cite{zhao2020diffaugment} applied to the sketches for training. We find that mild augmentation performs better in our scenario; in particular, we use translation for the augmentation. While augmentation does not necessarily improve the results when we use 30 input sketches generated from the Photosketch network $\Fnet$, we find that it is essential for model training with one or a few hand-drawn sketch inputs. More details are in \refsec{quantitative_eval}.}

\section{Experiments}
\subsection{Evaluations}
\label{sec:quantitative_eval}

\myparagraph{Datasets.}
\label{sec:dataset}
To enable large-scale quantitative evaluation, we construct a dataset of model sketching scenarios with ground truth target distributions defined as follows.  We transform the images from LSUN~\cite{yu2015lsun} horses, cats, and churches to sketches using PhotoSketch~\cite{li2019photo}, and hand-select sets of 30 sketches with similar shapes and poses to designate as the user input, as shown in \reffig{photosketch_result}.  To define the target distribution, \camready{we hand-select additional 2,500 images that match the input sketches. We select them out of 10,000 candidate images with the smallest chamfer distances~\cite{barrow1977parametric} to the designated inputs. (Candidate images retrieved} by the SBIR method of Bui~\etal~\cite{bui2017compact} was also considered but did not match poses as faithfully.)  Our method is given access only to the 30 designated sketches; the sets of \camready{2,500} real images represent the real but unseen target distributions.
\begin{figure*}
    \centering
    \includegraphics[width=1.\linewidth]{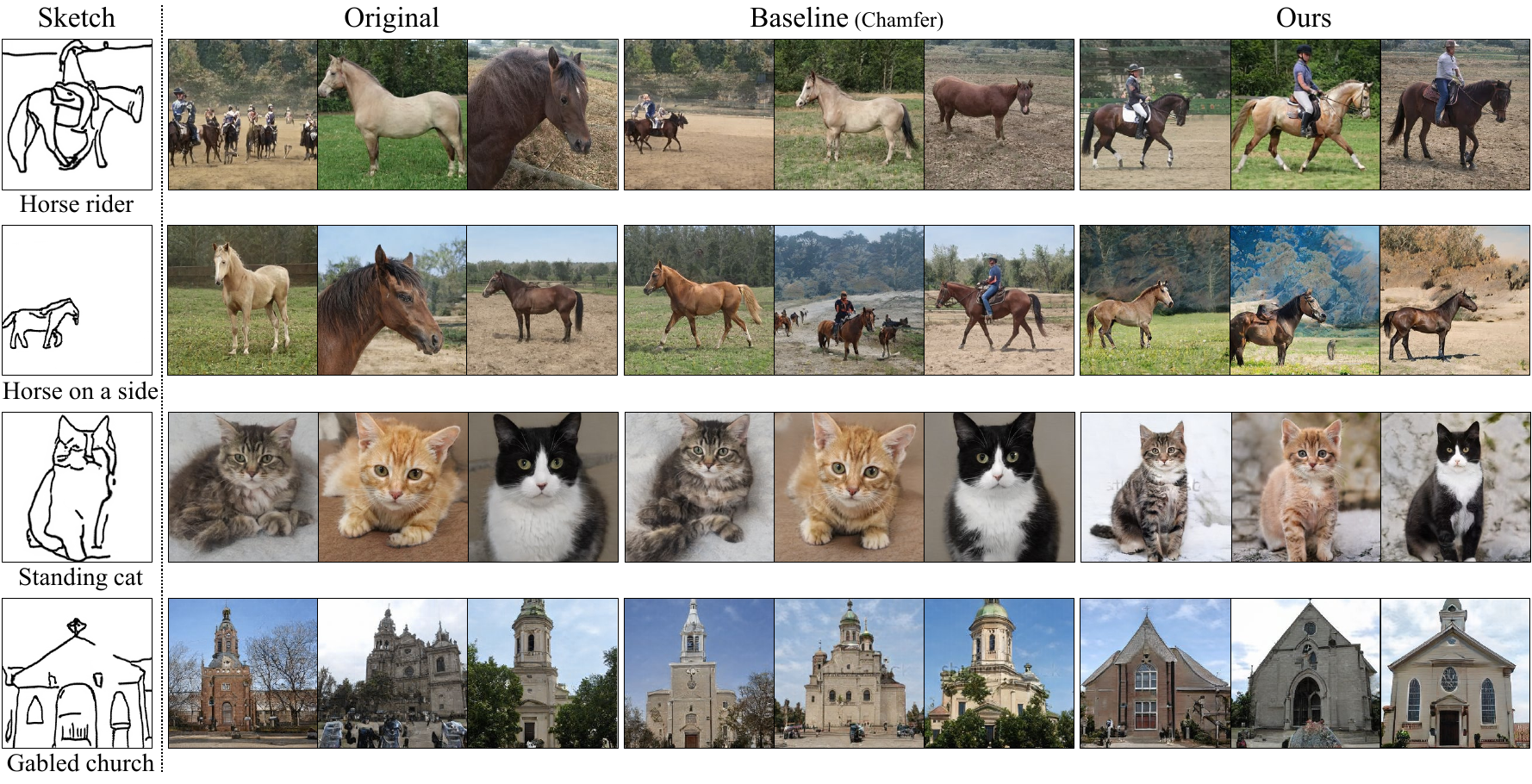}
           \vspace{-10pt}
    \caption{\textbf{Qualitative results on synthetic sketches.} Each row shows one model sketching task. For each of the four tasks, 30 PhotoSketch sketches are used for training; one training sketch is shown. We show samples generated from the original models, the models fine-tuned using \textbf{Baseline(Chamfer)} and the models customized using our method (\textbf{Full (w/o aug.)}), and the \textbf{Horse rider} model is \textbf{Full (w/ aug.)}). For each task, we show samples generated from the same three $z$. Following Karras~\etal~\cite{karras2020analyzing}, truncation $\psi=0.5$ is applied at all samples. We observe that the samples generated by our methods match the sketches better than the baseline.}
    \lblfig{photosketch_result}
        \vspace{-10pt}
\end{figure*} 

To test our method in a realistic scenario, we collect human sketches from the Quickdraw dataset~\cite{cheema2012quickdraw}. It is challenging to curate an evaluation set with human sketches. Hence, we evaluate on the test cases qualitatively.

\myparagraph{Performance metrics.}
\camready{We evaluate our models based on the \fid~\cite{heusel2017gans} (FID) between the generated images and the evaluation set. The FID measures the distribution similarity between the two sets, and serves as a metric for the diversity and quality of the generated images, as well as how well the images match the sketches. For a fair comparison, we evaluate each model by selecting the iteration with the best FID.}

\newcolumntype{N}{>{\centering\arraybackslash\hsize=.09\hsize}X}
\begin{table}[t]
{\small
    \centering
    \resizebox{1.\linewidth}{!}{
\begin{tabular}{cccccccc}
\toprule
\multirow{3}{*}{Family} & \multirow{3}{*}{Name} & \multicolumn{2}{c}{Training settings} & \multicolumn{4}{c}{FID $\downarrow$} \\ \cline{3-8} 
 &  & \multirow{2}{*}{\begin{tabular}[c]{@{}c@{}}No.\\ Samples\end{tabular}} & \multirow{2}{*}{Aug.} & \multirow{2}{*}{\begin{tabular}[c]{@{}c@{}}Horse\\ rider\end{tabular}} & \multirow{2}{*}{\begin{tabular}[c]{@{}c@{}}Horse on\\ a side\end{tabular}} & \multirow{2}{*}{\begin{tabular}[c]{@{}c@{}}Standing\\ cat\end{tabular}} & \multirow{2}{*}{\begin{tabular}[c]{@{}c@{}}Gabled\\ church\end{tabular}} \\
 &  &  &  &  &  &  &  \\ \midrule
Pre-trained & Original & N/A &  & \gray{50.43} & \gray{42.24} & \gray{58.71} & \gray{32.64} \\ \midrule
\multirow{2}{*}{Baseline} & Bui~\etal~\cite{bui2017compact} & 30 &  & 46.00 & 43.52 & 59.86 & 29.94 \\
 & Chamfer & 30 &  & 47.04 & 48.18 & 54.04 & 19.71 \\ \midrule
\multirow{6}{*}{Ours} & 1-sample & 1 &  & 29.25 & 41.50 & 44.68 & 26.88 \\
 & 5-sample & 5 &  & 33.11 & 41.61 & \textbf{31.20} & 23.28 \\
 \cdashline{2-8}
 & D scratch\textsuperscript{$\ddagger$} & 30 &  & 44.91 & 27.84 & 47.69 & 24.41 \\
 & W-shift\textsuperscript{$\ddagger$} & 30 &  & 30.66 & 34.86 & 42.24 & 17.88 \\
 & Full (w/o aug.) & 30 &  & 27.50 & \textbf{29.62} & 33.94 & \textbf{16.70} \\
 & Full (w/ aug.)  & 30 & \checkmark & \textbf{19.94} & 30.39 & 36.73 & 21.35 \\ \bottomrule
\end{tabular}
}
}
\caption{{\bf Quantitative analysis.} We report the \fid (FID) of the original models, baselines and our methods on four different test cases with synthetic sketch inputs. The details of the baselines are in \refsec{quantitative_eval}. We test on model variants trained on fewer training samples (No. Sample) and ablate our method by altering the training components. $\checkmark$ indicates translation augmentation is applied. $\uparrow$, $\downarrow$ indicate if higher or lower is better. Evaluations on the original models are in \gray{gray}, and the best value is highlighted in {\bf black}. ($\ddagger$: ``D scratch'' indicates that the sketch discriminator $\sketchD$ is initialized randomly; ``W-shift'' indicates that a shift in $\mathcal{W}$ space is the only tunable parameter of the generator.) }
\label{tbl:mainresults}
\end{table}

\begin{table}[t]
\centering
\resizebox{\linewidth}{!}{
\begin{tabular}{lcccc}
\toprule
 & \multicolumn{4}{c}{FID $\downarrow$} \\ \cmidrule(lr){2-5} 
 & Horse rider & Horse on a side & Standing cat & Gabled church \\ \midrule
Original model & \gray{50.43} & \gray{42.24} & \gray{58.71} & \gray{32.64} \\ \hdashline
$\losssketch$ & 27.39 & 40.65 & 50.09 & 19.33 \\
$\losssketch$+aug. & 28.28 & 39.03 & 49.52 & 21.60 \\
$\losssketch$+$\lossweight$ & 30.94 & 38.55 & 49.76 & 17.55 \\
$\losssketch$+$\lossweight$+aug. & 21.99 & 35.44 & 48.84 & 22.41 \\ \hdashline
$\losssketch$+$\lossimage$ & 27.50 & \textbf{29.62} & \textbf{33.94} & \textbf{16.70} \\
$\losssketch$+$\lossimage$+aug. & \textbf{19.94} & 30.39 & 36.73 & 21.35 \\ \bottomrule
\end{tabular}
}
\vspace{5pt}
\caption{\textbf{Ablation study.} We evaluate the effect of each component of our losses and data augmentation on four test cases with synthetic sketch inputs, and report the \fid (FID) tested on our evaluation set. Lower value is better. ``Original model'' denotes the pre-trained models (labelled in \gray{gray}).}
\label{tbl:loss_ablation}
\vspace{-10pt}
\end{table}

\myparagraph{Baselines.}
We compare our method to the following baselines.  We evaluate the effect on the model output when it is customized by shifting the latent $w_\text{new} = w + \Delta w$ using a constant vector $\Delta w$ derived by averaging samples that resemble the user sketch, similar to the vector arithmetic method proposed in Radford~\etal~\cite{radford2015unsupervised}:
$\Delta w =  \mathbb{E}_\text{match}[w] -  \mathbb{E}_\text{orig}[w]$.
Here $ \mathbb{E}_\text{orig}[w]$ is the mean latent in the original unmodified model and $ \mathbb{E}_\text{match}[w]$ is the mean latent over a subset of samples where the image resembles the user sketch according to a simple criterion.  We implement two baseline variants, which use different methods to sample  $ \mathbb{E}_\text{match}[w]$.  \textbf{(1) Baseline (SBIR):} selects best matched samples using the sketch-based image retrieval method by Bui~\etal~\cite{bui2017compact}, \textbf{(2) Baseline (Chamfer):} matches samples using the symmetric Chamfer distance $d(x, y) + d(y, x)$ between the input sketch $y$ and a sketch of the image $x$ as computed by PhotoSketch~\cite{li2019photo}.  To estimate $ \mathbb{E}_\text{match}[w]$, we take the top-matched 10,000 images from 1 million samples, and we score each image using the minimum distance to all user sketches.
\begin{figure*}
    \centering
    \includegraphics[width=\linewidth]{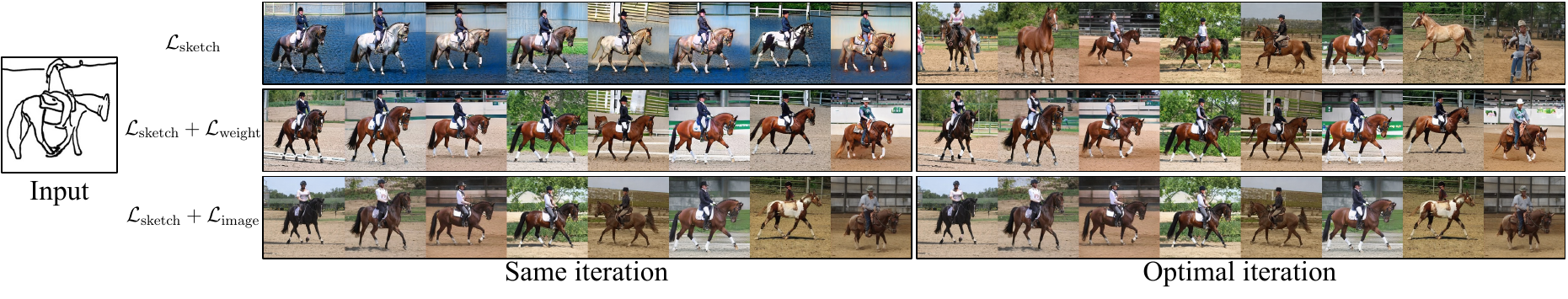}
    \caption{\textbf{Effect of regularization methods.} We compare models trained with (top) only $\losssketch$, with (middle) weight regularization $\lossweight$, and with (bottom) image regularization $\lossimage$. \textbf{(Left)} the snapshots at the same iteration. \textbf{(Right)} the best iterations selected based on our evaluation metric. We observe that the images look more realistic with either regularization applied, and the model trained with image regularization obtains better diversity than the one trained with weight regularization.}
    \lblfig{reg}
    \vspace{-8pt}
\end{figure*}

 \reftbl{mainresults} shows the quantitative comparisons. \camready{We note that our method obtains significantly better FID than \textbf{Baseline (SBIR)} and \textbf{Baseline (Chamfer)}. The result agrees with our qualitative comparison in \reffig{photosketch_result}, where the baseline does not match user sketches nearly as well as our method (\textbf{Baseline (Chamfer)} is chosen for visual comparison because it obtains better average FID than \textbf{Baseline (SBIR)}). }

\myparagraph{Ablation studies.}
\camready{We first investigate the effects of our regularization methods and data augmentation}. The results are shown in \reftbl{loss_ablation}.

\newcommand{\rarrow}{$\rightarrow$}
\noindent\textbf{Augmentation.} 
\camready{We find that augmentation does not necessarily improve the performance for sketches generated from Photosketch. With image regularization applied, the \textbf{horse rider} model benefits from augmentation, whereas \textbf{horse on a side}, \textbf{standing cat} and \textbf{gabled church} models perform better without augmentation. However, we find that augmentation is critical when the training inputs are human-created sketches, as discussed later}.%

\noindent\textbf{Comparing regularization methods.} \camready{Either regularization method $\lossimage$ or $\lossweight$ improves FID over models trained with only $\losssketch$, while models trained with image regularization $\lossimage$ outperform models trained with $\lossweight$. This agrees with our observation in \reffig{reg}, %
which shows snapshots of models trained with and without regularization. At the same training iteration, both regularization methods preserve image quality, and image regularization obtains the most diverse outputs. When the best iterations are selected for each method, the comparison again reveals that the regularized models obtain a better balance of shape matching and image quality.}

We also investigate the effects of other training components, as shown in \reftbl{mainresults}. \camready{The following analysis focus on models trained without augmentation, since augmentation in general is not beneficial to synthetic sketch inputs. We refer \textbf{$\losssketch$+$\lossimage$} as \textbf{Full (w/o aug.)} and \textbf{$\losssketch$+$\lossimage$+aug.} as \textbf{Full (w/ aug.)}.}

\noindent\textbf{D scratch.} To test whether using pre-trained weights for the sketch discriminator $\sketchD$ is necessary, we evaluate a variant with $\sketchD$ initialized randomly. We observe a huge decrease in performance \camready{in most cases}.  This indicates that pre-training in discriminators is important even when switching the training domain from images to sketches. The finding is consistent with prior work on few-shot GAN finetuning~\cite{mo2020freeze,li2020fewshot}.

\noindent\textbf{W-shift.} We test on a variant where the only tunable parameter is a bias added to the mapping network, which effectively performs a shift in the $\mathcal{W}$ space. 
\camready{We observe this method leads to a reasonable performance, despite being worse than our full method by a margin. This shows that our training procedure can potentially serve as a flexible latent discovery method. 
However, in general, tuning the entire mapping network makes our method more effective.}

\myparagraph{Fewer sketch samples.}
We test if our method is able to work on a smaller number of sketches. For each task, we trained models with only 1 or 5 sketches, selected from the previous 30 sketches. The results are reported in \reftbl{mainresults}.

\camready{We observe that the models trained with 1 or 5 sketches improve upon the original model. In most cases, training on 30 sketches still outperforms, which shows the degree to which results can be improved when the user provides more sketches. In the \textbf{standing cat} task, training with 5 sketches achieves slightly better results, and in the \textbf{horse on a side} task the improvement is small. The other two tasks improves significantly when 30 sketches are used. Training a model with very few sketches remains challenging, but our method improves with the number of user sketches.}%

\begin{figure}
    \centering
    \begin{tabular}{@{}c@{}}
    \includegraphics[width=\linewidth]{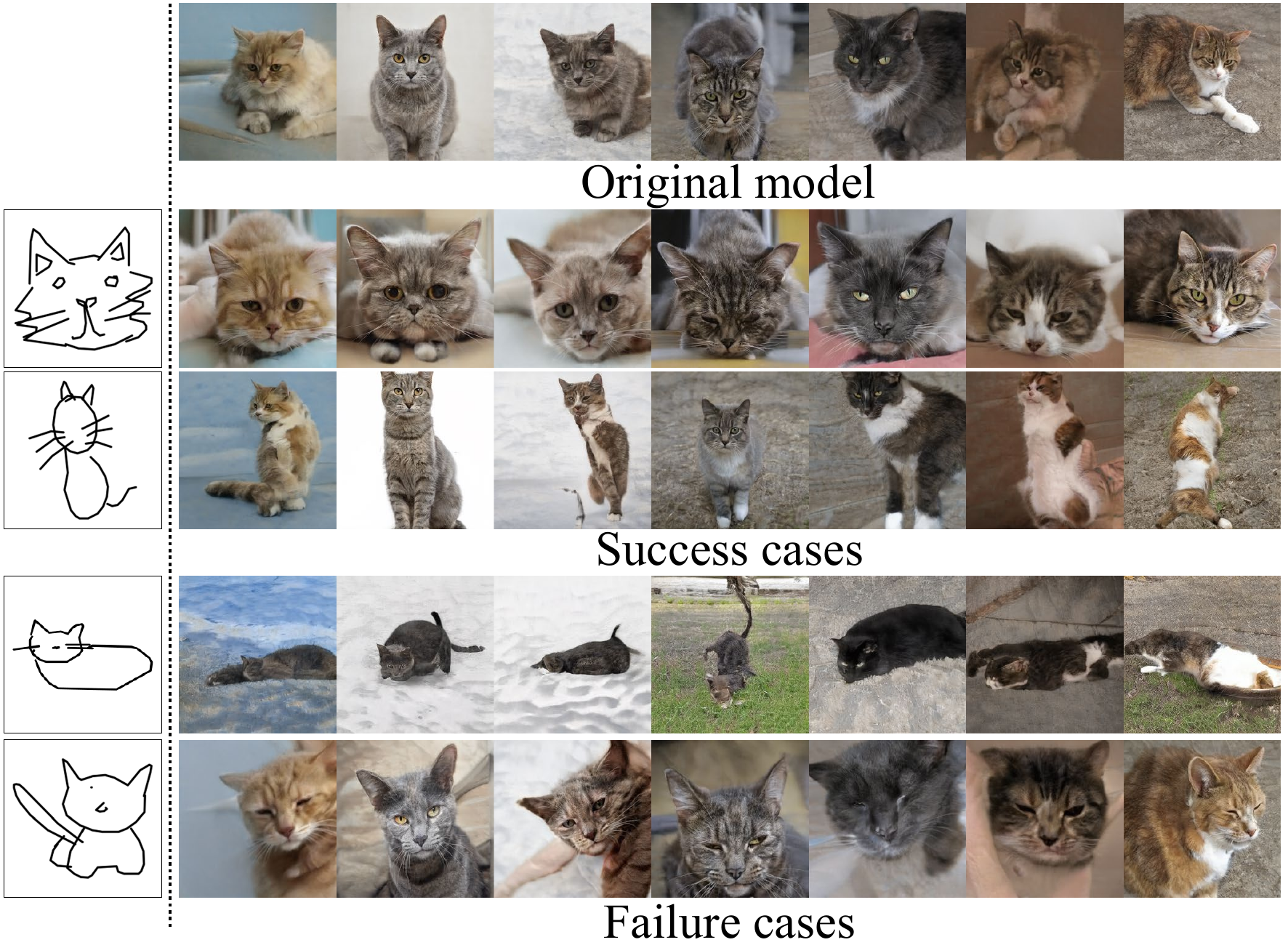}  \\
    \includegraphics[width=\linewidth]{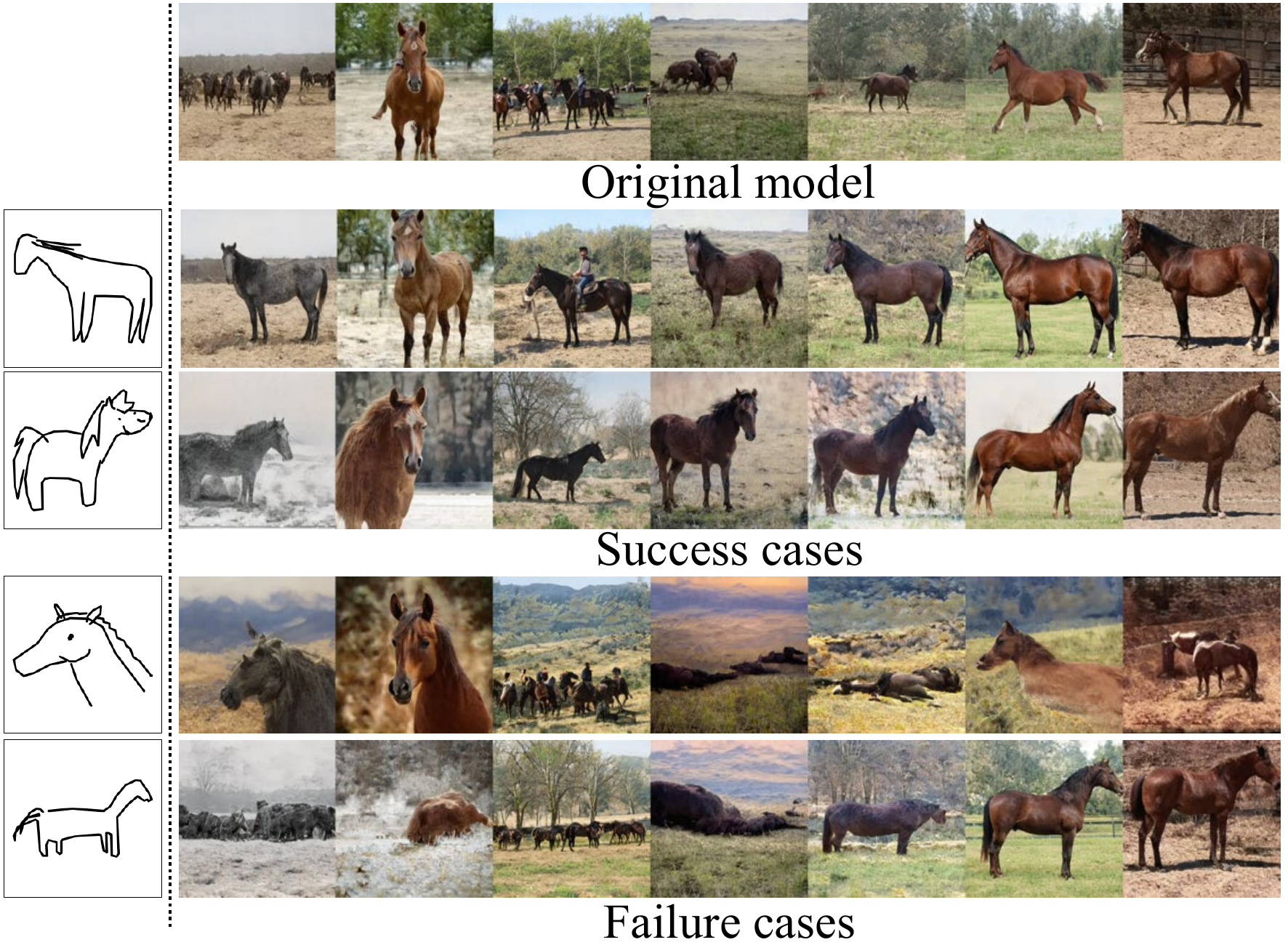} 
    \end{tabular}
    \caption{\textbf{Model creation using a single human-created sketch.} Each row shows uncurated samples generated from a model trained on a single sketch from Quickdraw~\cite{cheema2012quickdraw}. Same noise $z$ is used and truncation $\psi=0.5$ is applied to each model. Results on cats \textbf{(top)} and horses \textbf{(bottom)} are shown. For each category, the 1st row is the original pre-traiend model, 2nd and 3rd rows represent the success cases, and the last two rows are the failure cases.}
    \lblfig{quickdraw_single}
\end{figure}

\begin{figure}
    \centering
    \includegraphics[width=\linewidth]{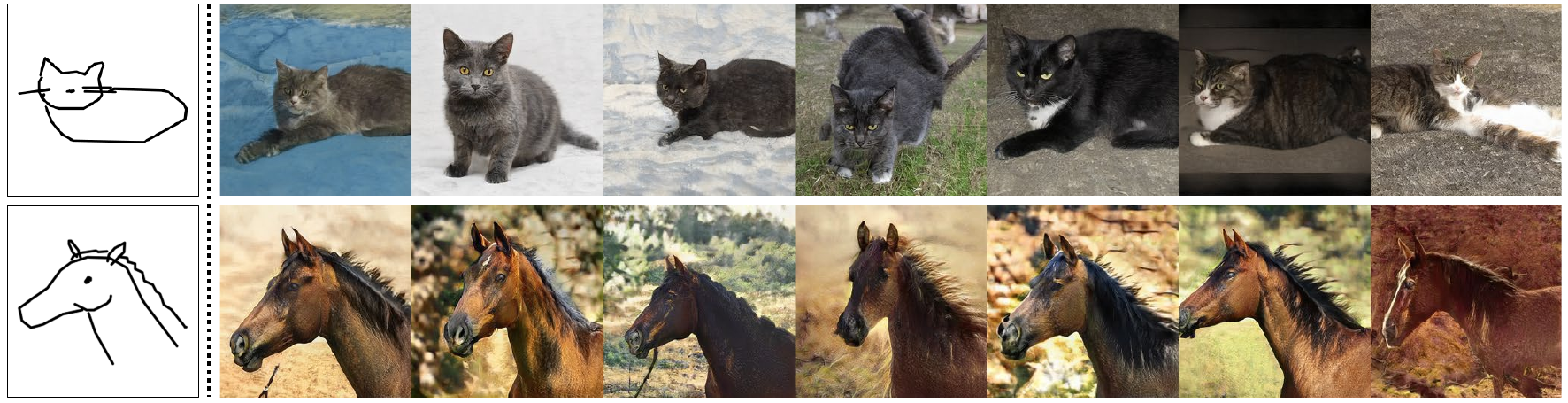}
    \caption{\textbf{Model creation using multiple human-created sketches.} Some failure cases in \reffig{quickdraw_single} can be improved with more input sketches. 3 similar sketches are used in the cat model \textbf{(top)} and 4 are used in the horse model \textbf{(bottom)}. Samples are generated in the same fashion as in \reffig{quickdraw_single}}
    \lblfig{quickdraw_more}
\end{figure}

\begin{figure}
    \centering
    \includegraphics[width=\linewidth]{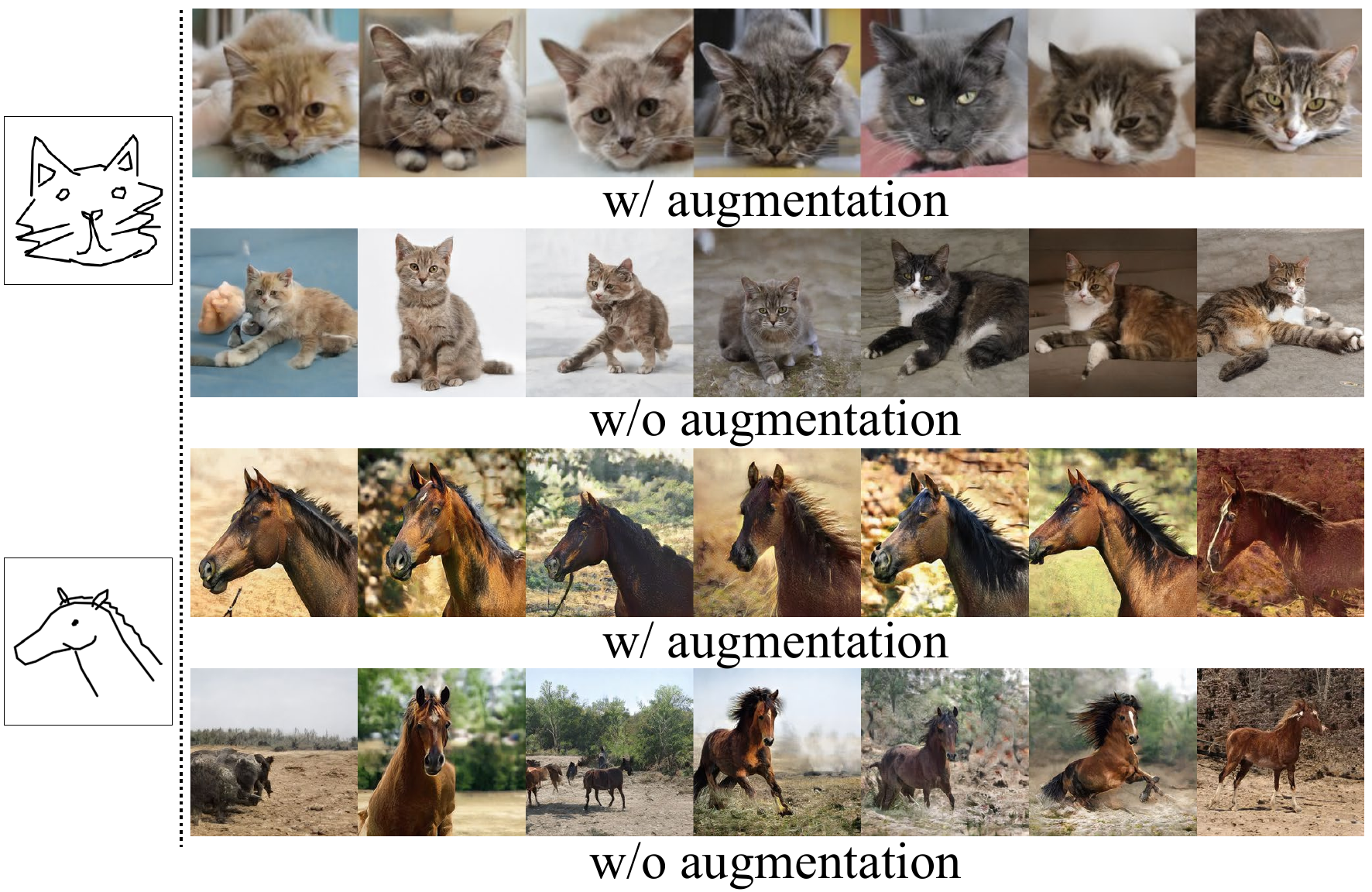}
    \caption{\textbf{Effect of augmentation on human-created sketches.} We take models from \reffig{quickdraw_single} (top) and \reffig{quickdraw_more} (bottom), and observe that only models trained with augmentation generate images that faithfully match the sketches.}
    \lblfig{quickdraw_aug}
\end{figure}

\begin{figure}
    \centering
    \includegraphics[width=0.8\linewidth]{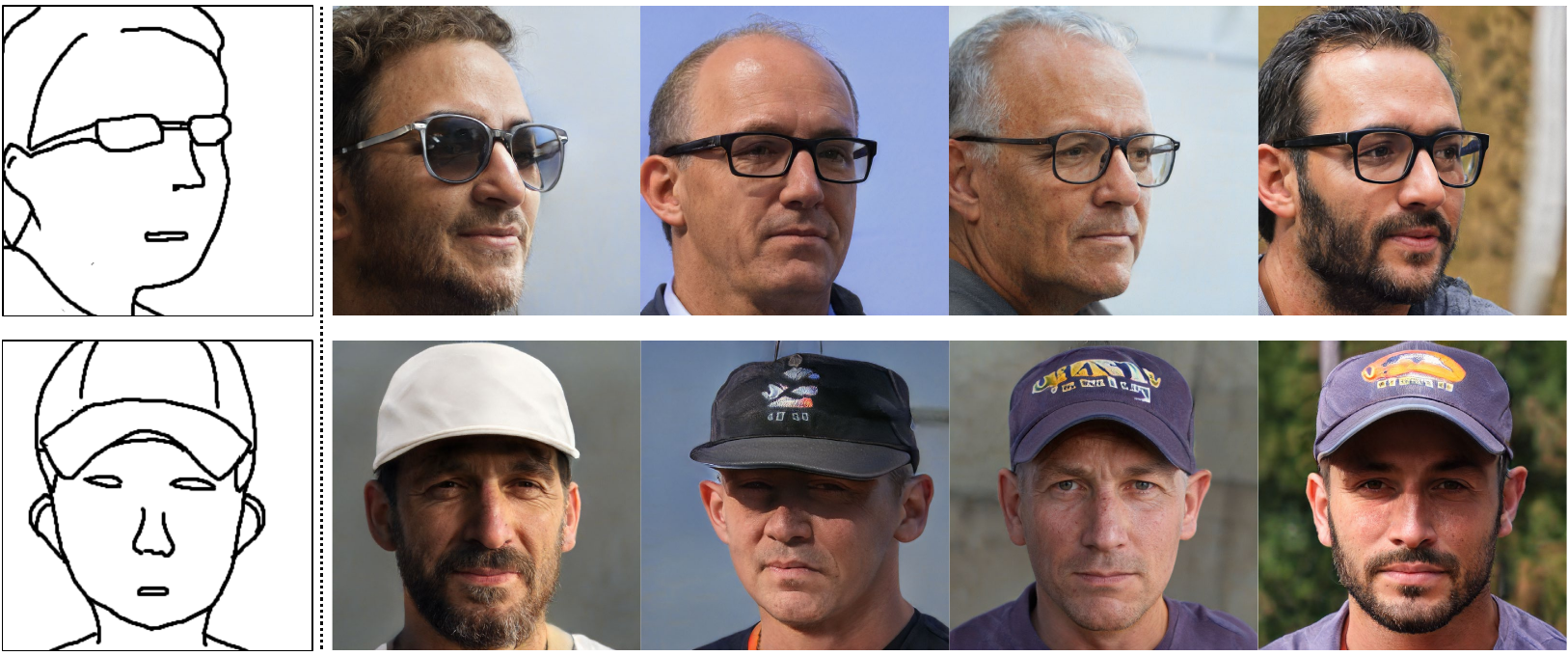}
    \caption{\textbf{Customizing FFHQ models.} Each row shows samples on a customized FFHQ model, trained on 4 human-created sketches (1 is shown).}
    \label{fig:ffhq}
\end{figure} \myparagraph{Testing using real human sketches.}
To make GAN customization available to everyday users, we then test the effectiveness of our method on hand-drawn sketches from novice users. We collect cat and horse sketches from the Quickdraw~\cite{cheema2012quickdraw} dataset as training images. \camready{We first train models on a single sketch, and show success and failure cases in \reffig{quickdraw_single}. We note that the image to sketch translation network Photosketch~\cite{li2019photo} used in our method is trained on a sketch style that is different from the Quickdraw dataset. Despite the difference in the styles, our method succeeds on sketches that are more contour-like and depict simple poses (e.g., cat head). However, there is headroom for improvement, to support a wider range of sketching styles and poses. In particular, the results are worse with sketches that have more abstract styles or complex poses. We also observe that performance on difficult cases can be improved by increasing the number of input user sketches, as shown in \reffig{quickdraw_more}. Also, we find that augmentation is essential for our method to be successful for user sketches. As shown in \reffig{quickdraw_aug}, given the same input sketches, only the model trained with augmentation will faithfully generate images that match the sketches.}

\camready{We apply our method to generative model of faces as well. We customize the StyleGAN2 FFHQ models~\cite{karras2020analyzing} with 4 human-drawn sketches using our \textbf{Full (w/ aug.)} method, \camrdy{and we find that using $\lambda_\text{image}=0.5$ produces better results in this case.} Results are shown in Fig.~\ref{fig:ffhq}, and one can observe the generated output resembles the input sketches.}

\subsection{Applications}
\label{sec:realworld}
In this section, we discuss several ways of applying our method to image editing and synthesis tasks. We show that with our customized models, one can perform latent space edits and manipulate natural images.
We also demonstrate that it is possible to interpolate between the customized models.

\begin{figure}
    \centering
    \includegraphics[width=1.0\linewidth]{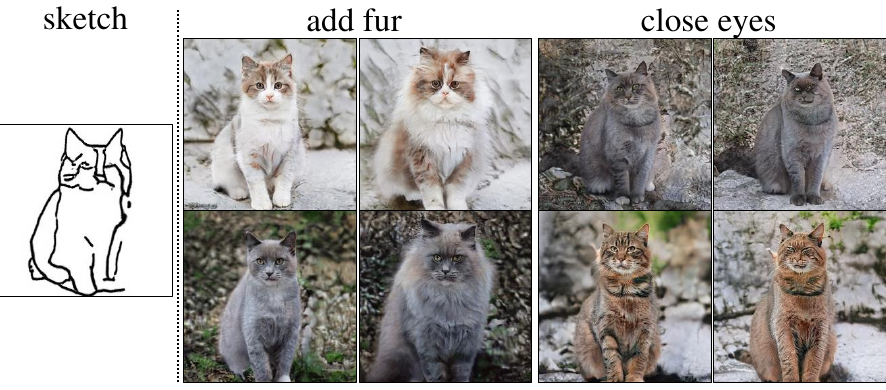}
           \vspace{-10pt}
    \caption{\textbf{Editability of customized models.} We apply the edits reported in H{\"a}rk{\"o}nen~\etal~\cite{harkonen2020ganspace} to the customized models. We observe that the edit operations discovered from the original model gives the same effect to the customized models}
    \lblfig{control}
\vspace{-10pt}
\end{figure} %
\begin{figure}
    \centering
    \includegraphics[width=1.0\linewidth]{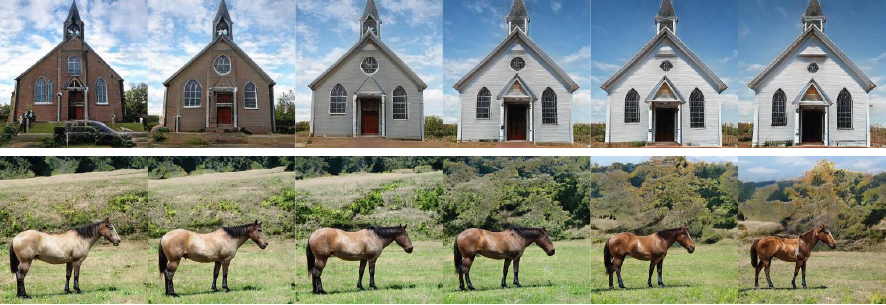}    
    \caption{\textbf{Interpolation using customized models.} Latent space interpolation is smooth with our customized models. Here we show \textbf{gabled church (top)} and the \textbf{horse on a side (bottom)} results.}
    \label{fig:interp}
    \vspace{-5pt}
\end{figure} 

\myparagraph{Latent space edits.}
Interpretable user controls in a generative model are useful for many graphics applications. We investigate whether this property still holds for our customized models.
More importantly, the edit operations should be identical to the original models, to avoid the need to run latent discovery algorithms again on each customized model. 
To investigate this, we apply the latent discovery method GANSpace~\cite{harkonen2020ganspace} to the original models. By moving along the reported latent directions, we observe that our customized models can perform the exact same manipulations from the results in H{\"a}rk{\"o}nen~\etal, as shown in \reffig{control}.
Since we are only tuning the mapping network of the generator, our method did not change how the model is processing the $w$-space latents. As a result, properties on latent edits are preserved. In addition, we observe that latent interpolation remains smooth in our models, as shown in \reffig{interp}

\begin{figure}
    \centering
    \includegraphics[width=1.0\linewidth]{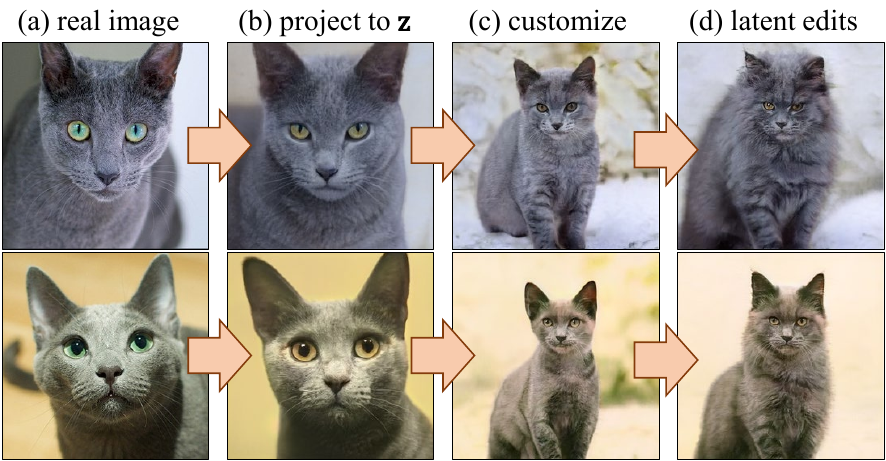}
             \vspace{-10pt}
    \caption{\textbf{Natural image editing with original and customized models.} Given a real image as input \textbf{(a)}, we project the image to the original model's $z$ using Huh~\etal~\cite{huh2020ganprojection} \textbf{(b)}. We then feed the projected $z$ to the \textbf{standing cat} model trained on sketches, which effectively edits input to match the sketches \textbf{(c)}. Furthermore, we showed the image can be further edited using GANSpace~\cite{harkonen2020ganspace} \textbf{(d)}. Similar to \reffig{control}, both images from the customized model can be applied with the ``add fur'' operation.}%
    \lblfig{natural}
           \vspace{-10pt}
\end{figure}

\myparagraph{Natural image editing with our models.}
Our method is capable of ``editing'' an original model to create a new model that matches user sketches, but is it possible to edit a single photograph using our new model? We show that natural image editing can be realized by image projection. To illustrate, we project the natural image to a noise $z$ from the original model using Huh~\etal~\cite{huh2020ganprojection}. Since our method modifies the shape and pose while preserving \camready{the texture of background and objects} under the same $z$, we can feed the projected $z$ to the customized model, and the output is effectively the transformed version of the input image with a new shape and pose matching the sketch. We also verified that the latent space edits (\reffig{control}) still apply to the ``transformed'' photograph. The results are shown in \reffig{natural}.

\begin{figure*}[t]
    \centering
    \includegraphics[width=0.95\linewidth]{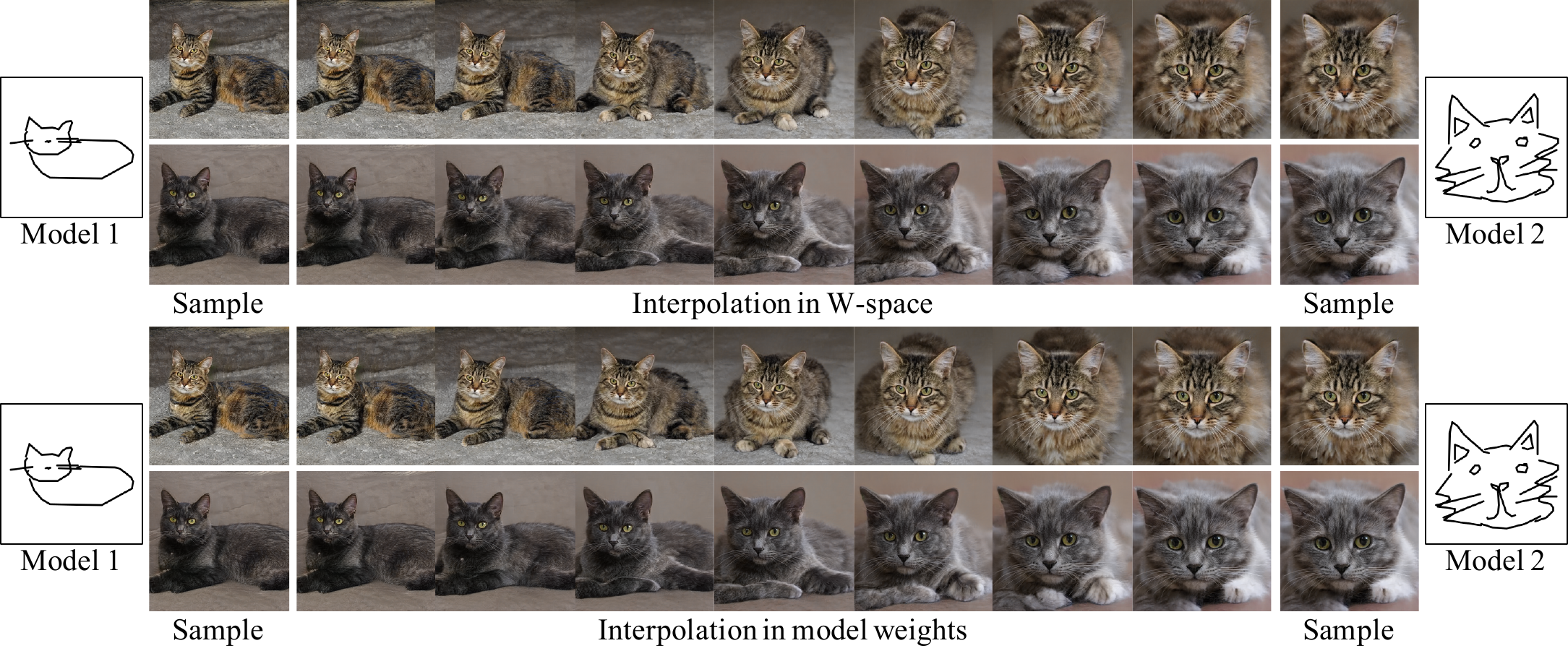}
    \caption{\textbf{Interpolating between customized models.} We can interpolate between the customized model by interpolating (top) the W-latents or (bottom) the model weights.  We show image samples from model 1 and 2 on the left and right side, respectively, and the middle shows the interpolation results. Model 1 and 2 are from \reffig{quickdraw_more} and \reffig{quickdraw_single}, respectively.}
    \label{fig:model_interp}
\end{figure*}

\myparagraph{Interpolating between customized models.}
After models are customized, we demonstrate that it is possible to interpolate between these models in two ways, as shown in \reffig{model_interp}. (1) First, we feed the same noise vector $z$ to two different models to obtain two latents $w_1$, $w_2$ in the $\mathcal{W}$ space. We then feed $(1-\alpha)w_1 + \alpha w_2$ to the synthesis network to obtain images with interpolated shapes between the two models, where $\alpha \in [0,1]$ controls the interpolation. (2) Also, we observe that the same effects can be achieved by directly interpolating the model weights $\theta_1$, $\theta_2$. In particular, the interpolated models have weights $(1-\alpha)\theta_1 + \alpha \theta_2$. We note that the similar task has been explored by concurrent work~\cite{avrahami2021gan}.

\begin{figure}
    \centering
    \includegraphics[width=1.0\linewidth]{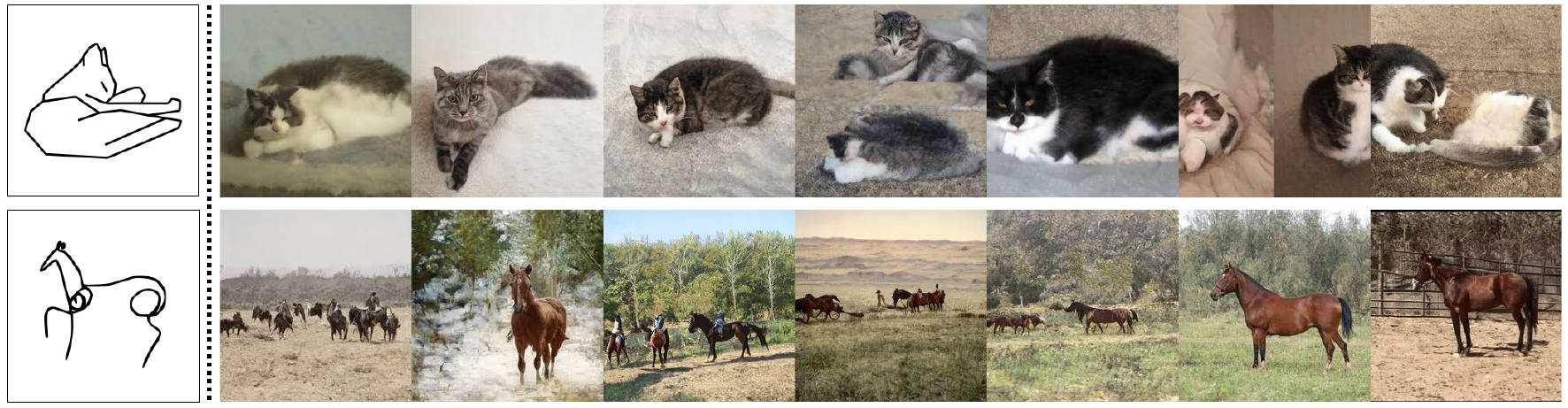}
        \vspace{-10pt}
    \caption{\textbf{Failure cases.}  Our method is not capable of generating images to match the Attneave's cat sketch~\cite{Attneave1954Some} \textbf{(top)} or the horse sketch by Picasso \textbf{(bottom)}.}
    \lblfig{failure}
\end{figure}

\section{Discussion}
In this work, we present a method that enables the user creation of customized generative models, by leveraging off-the-shelf pre-trained models and cross-domain training. The required input of our method is just one or more hand-drawn sketches, which makes model creation possible for novice users. Our method overcomes the large domain gap between user sketches and generator parameter space, and it is capable of generalizing to sketches in different styles.

However, plenty of improvement remains for our method. We have shown our method generalizes to other sketch styles and different poses in \refsec{realworld}, but our method is not guaranteed to work for all sketches. \camready{For example, we tested our method on horse sketches from Picasso and the Attneave's sleeping cat~\cite{Attneave1954Some}. As shown in \reffig{failure}, our method cannot create a model that matches the pose faithfully. We note that Picasso's sketches are drawn with a distinctive style, and Attneave's cat depicts a complex pose, both of which are potential reasons for the failure.}
Another limitation is that customizing a model in real-time is impossible with our current method, since our models take more than 30K iterations to train. 
Our method requires access to the training set of the original model, which may make it inappropriate for settings where this data is not available. \camready{Finally, while our method enables flexible control of shape and poses, it cannot customize other properties such as color and texture. To expand expressiveness, we note our cross-domain loss can be applied to other inputs such
as VGG features, color scribbles, or semantic layouts.}

\myparagraph{Acknowledgment.} 
We thank Nupur Kumari and Yufei Ye for proof-reading the drafts. We are also grateful for helpful discussions from Gaurav Parmar, Kangle Deng, Nupur Kumari,  Andrew Liu, Richard Zhang, and Eli Shechtman.
We are grateful for the support of Sony Corporation, Naver Corporation,  DARPA SAIL-ON HR0011-20-C-0022 (to DB), and Signify Lighting Research.

{\small
\bibliographystyle{ieee_fullname}
\bibliography{main}
}

\setcounter{section}{0}
\renewcommand\thesection{\Alph{section}}
\renewcommand{\thefootnote}{\arabic{footnote}}

\clearpage
\noindent{\Large\bf Appendix}
\vspace{5pt}

\noindent Below we include additional implementation details as well as extra results. 

\section{Implementation Details.}
\myparagraph{Training details}
\label{sec:impl_details}
We use the same training hyper-parameters as~\cite{karras2020analyzing}. In particular, we are using softplus for GAN loss, and R1 regularization~\cite{mescheder2018training} on both the sketch and image discriminator, $\sketchD$ and $\imageD$. We do not use path length regularization, as it has no effect on the latent mapping network. %
\camready{Also, we set the batch size to 4 for all of our experiments, except when the sketch inputs are less than four, where we set the batch size to 1.}

\myparagraph{Hyperparameters.}
\camready{We use the same hyperparameters for our \textbf{full} method in all of our experiments. In particular, we use $\lambda_{\text{image}}=0.7$}

\camready{In \refsec{quantitative_eval}, we compared several variants of our method in our ablation studies. To make the comparison fair, for each variant, we tuned the loss weights for optimal performance. In \reftbl{loss_weight},  we list the hyperparameters used for each variant. The only exception is that we use $\lambda_{\text{weight}}=50$ for the \textbf{$\losssketch$ + $\lossweight$} and \textbf{$\losssketch$ + $\lossweight$ + aug.} variant model trained on the \textbf{standing cat} task. Also, if the variants are not listed in the table, the same loss weights as the full method are used. 
The search space of the $\lambda_{\text{image}}$ is $[0.3, 0.5, 0.7, 1.0]$, and the search space of $\lambda_{\text{weight}}$ is $[0.1, 1, 10, 50, 100, 1000]$.}

\begin{figure*}
    \centering
    \includegraphics[width=.75\linewidth]{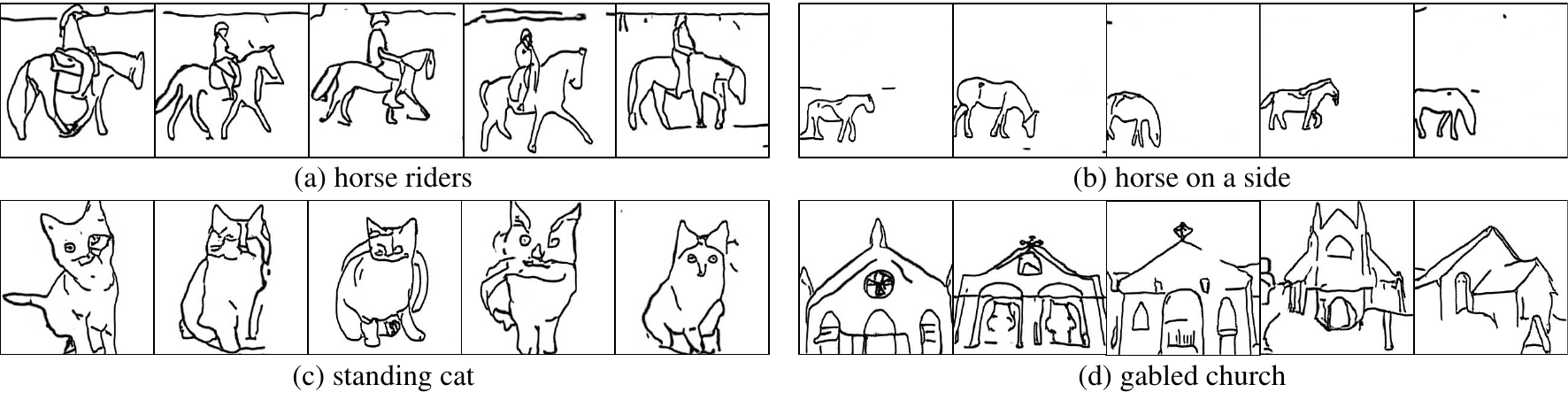}
        \vspace{-5pt}
    \caption{\textbf{Example of sketches used for training.} For each task (a, b, c, d), 30 sketches with similar shapes and layouts are hand-selected as training samples, where above shows subsets of 5 sketches.}
    \label{fig:exemplars}
        \vspace{-20pt}
\end{figure*}

\begin{figure}[h]
    \centering
    
    \includegraphics[width=1.\linewidth]{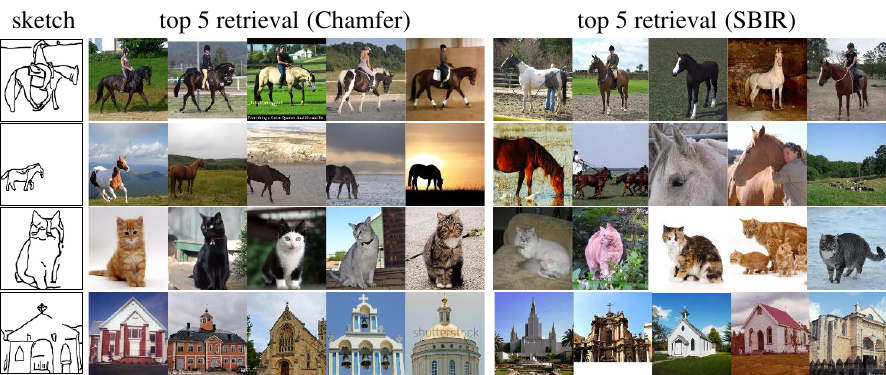}

    \caption{\textbf{Comparison between retrieval methods.} We compare retrieval methods between chamfer distance~\cite{barrow1977parametric} and SBIR method of Bui~\etal~\cite{bui2017compact}. We find that the retrievals using chamfer distance matches the input sketches better than those using Bui~\etal. Left shows the example query out of the 30 sketches used for the retrieval.}
    \label{fig:evalset}
\end{figure}

\begin{figure}[h]
    \centering
    \includegraphics[width=1.\linewidth]{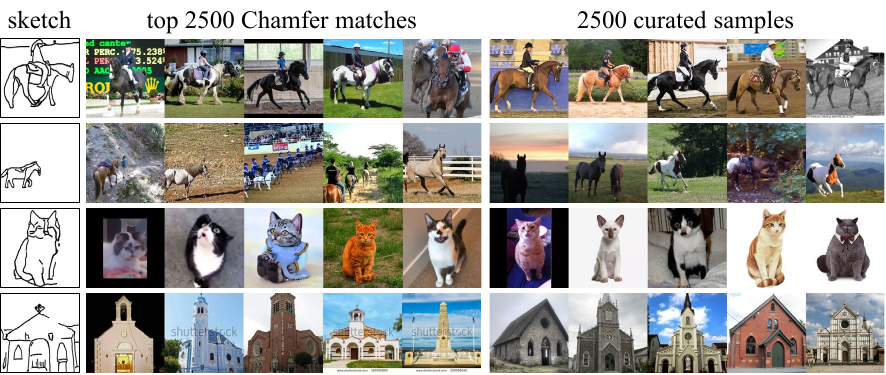}

    \caption{\textbf{Curated evaluation set.} We show random samples from the top 2,500 matches using chamfer distance~\cite{barrow1977parametric} \textbf{(left)} and 2,500 hand-selected images \textbf{(right)}. The quality of the evaluation set is improved after curation.}
    \label{fig:clean_eval}
\end{figure}

\myparagraph{Data collection.}
In \refsec{quantitative_eval}, we selected sets of
30 sketches with similar shapes and poses to designate as
the user input: examples of sketches from these sets are shown in \reffig{exemplars}.  To evaluate generation quality, we collected %
images that match the input sketches from LSUN~\cite{yu2015lsun}.  To retrieve matching images, we experimented with two sketch-image cross-domain matching methods. We applied both the SBIR method of Bui~\etal~\cite{bui2017compact} and chamfer distance~\cite{barrow1977parametric}. Both of these retrieval results are shown in \reffig{evalset}. 
We observe that with chamfer distance, the retrieved images match poses of the sketches more faithfully. As a result, we adopt this method to generate our evaluation sets. \camready{However, we notice that there still exists outliers after the retrieval; hence, we hand-selected 2,500 images out of top 10,000 matches to curate the evaluation sets. A comparison between curated dataset and top chamfer matches are shown in \reffig{clean_eval}.}

\myparagraph{Evaluation procedure.}
To evaluate each model, we sample \camready{2,500} images without truncation and save them into png files. Likewise, the evaluation set described in \refsec{quantitative_eval} consists of \camready{2,500} 256$\times$256 images stored in png. \camready{We evaluate the \fid values using the CleanFID code~\cite{parmar2021cleanfid}}.

\section{Additional results}
\label{sec:additional_results}
\paragraph{Other evaluation metrics.}
\camready{We report Perceptual Path Length (PPL)~\cite{karras2019style} in \reftbl{other_metrics}. We find that our method improves the original models' PPL, and beats the baselines. We note that our model focuses on fewer modes than the original one, so interpolations are smoother on average, leading to smaller PPL.}

\camready{In addition, Precision, and Recall metrics~\cite{kynkaanniemi2019improved} are reported in \reftbl{other_metrics}. The precision measures the proportion of generated samples that are close to the real dataset in VGG feature space~\cite{simonyan2014very}, and the recall measures the proportion of real dataset that are close to generated samples in VGG feature space. We note that models with better results often have higher precision and lower recall. We expect our method to increase precision as it refines the generated distribution to better match the target distribution.  But since our task aims at finding a subset of the source distribution, our method theoretically \textit{cannot} increase the recall: increasing recall would require synthesizing new modes of real data without access to any new real examples.  In our setting, the ideal maximizes precision while maintaining recall unchanged from the pre-trained model. A drop in recall reveals some loss in diversity, and measures headroom for improving upon our method.}

\begin{figure}[t]
    \centering
    \includegraphics[width=.82\linewidth]{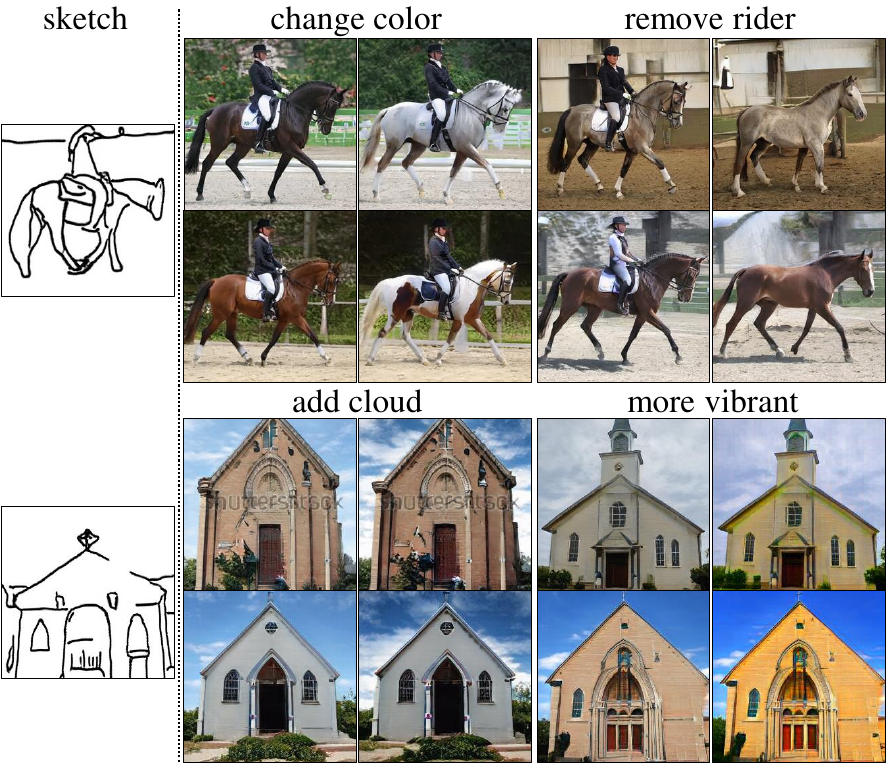}
    \caption{\textbf{Additional latent edit results.} Similar to Figure 7 of the main text, we show additional results of applying GANSpace~\cite{harkonen2020ganspace} edits to our customized models, \textbf{horse rider} (top) and \textbf{gabled church} (bottom).}
    \label{fig:more_edits}
\end{figure}

\begin{figure*}
    \centering
    \includegraphics[width=1.\linewidth]{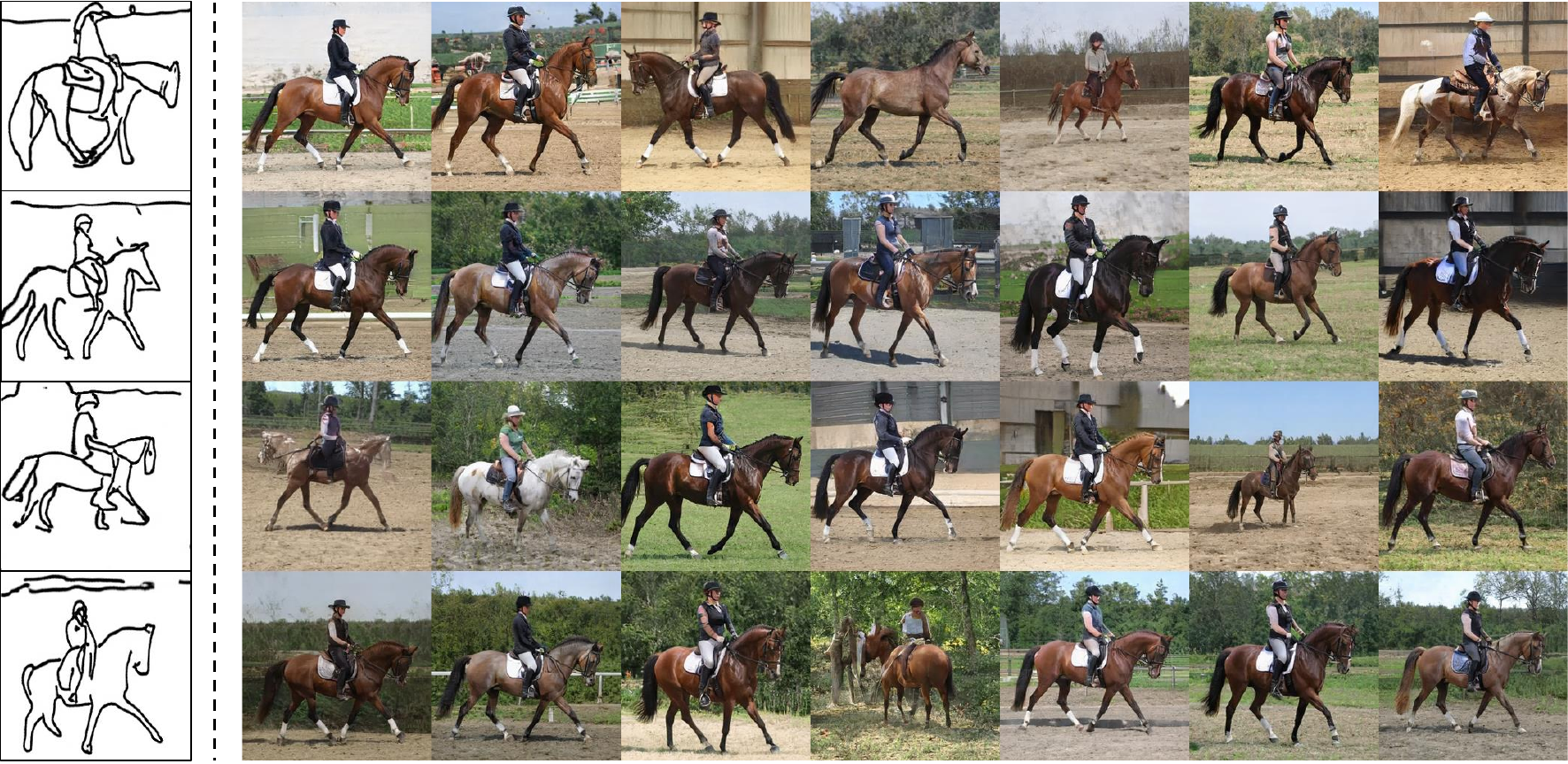}
    \caption{Uncurated samples of the \textbf{horse rider} model. Truncation $\psi=0.5$ is applied to generate the images.}
    \label{fig:horse_ride}
\end{figure*}

\begin{figure*}
    \centering
    \includegraphics[width=1.\linewidth]{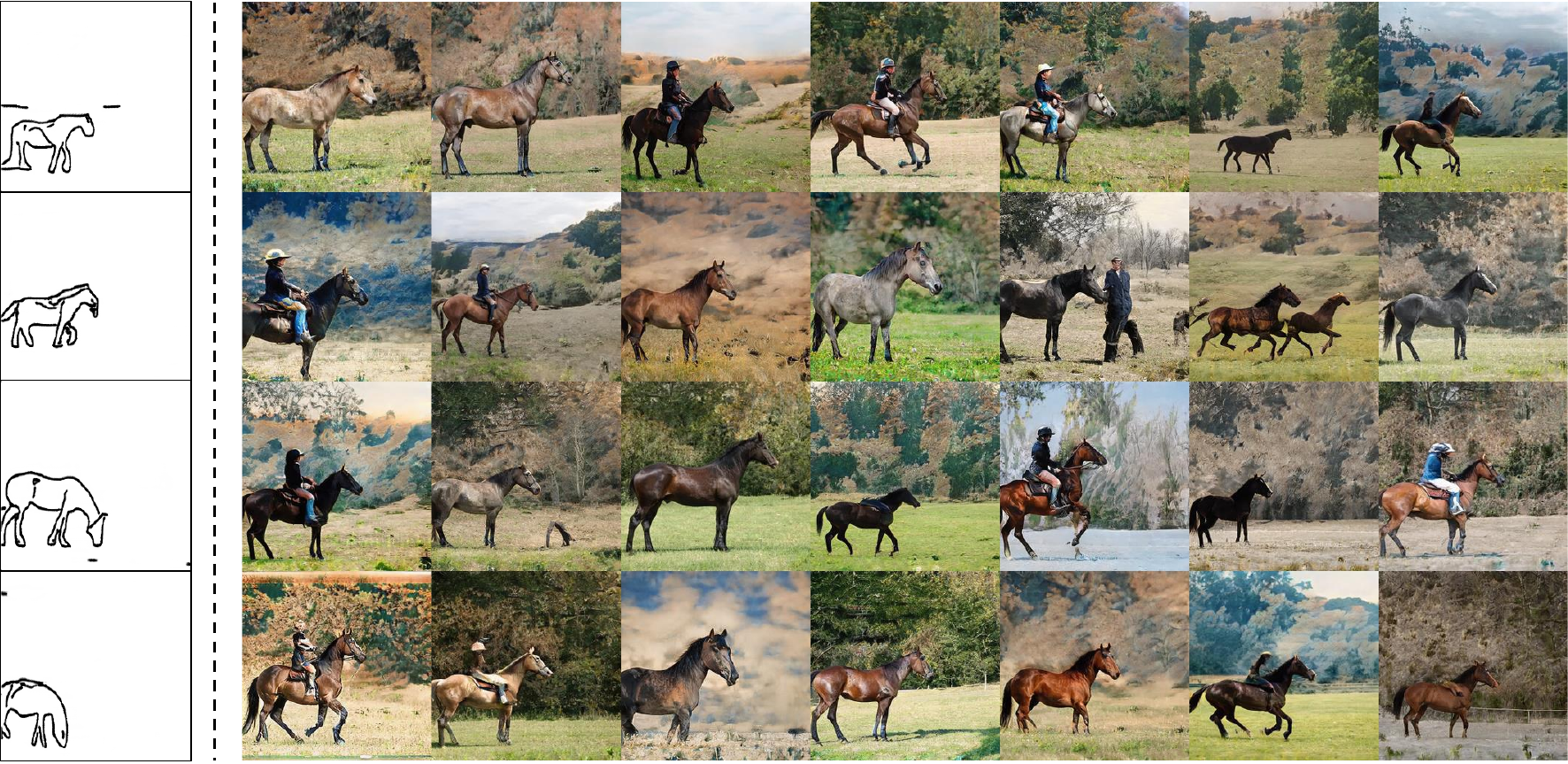}
    \caption{Uncurated samples of the \textbf{horse on a side} model. Truncation $\psi=0.5$ is applied to generate the images.}    \label{fig:horse_side}
\end{figure*}

\begin{figure*}
    \centering
    \includegraphics[width=1.\linewidth]{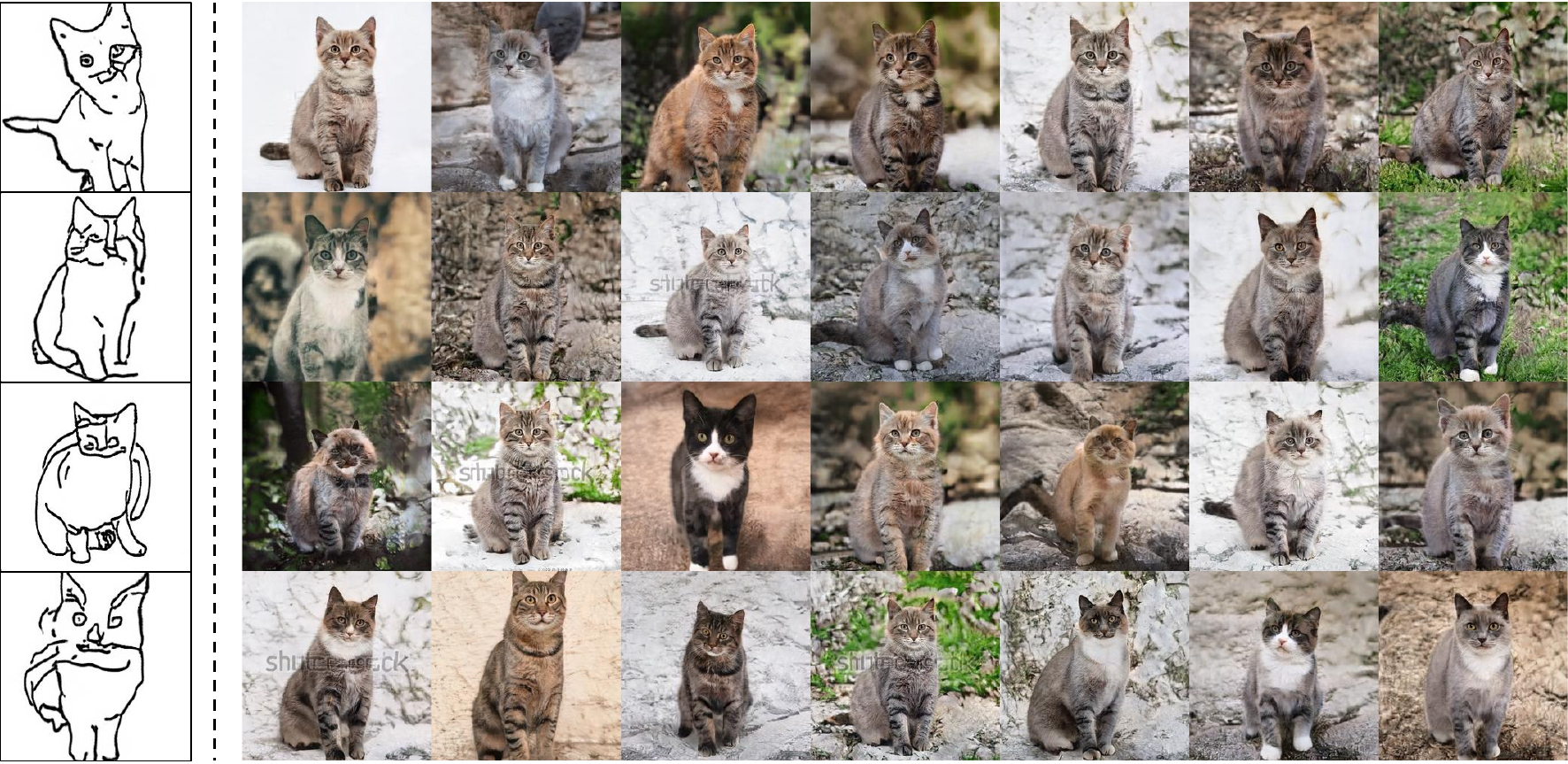}
    \caption{Uncurated samples of the \textbf{standing cat} model. Truncation $\psi=0.5$ is applied to generate the images.}    \label{fig:cat_stand}
\end{figure*}

\begin{figure*}
    \centering
    \includegraphics[width=1.\linewidth]{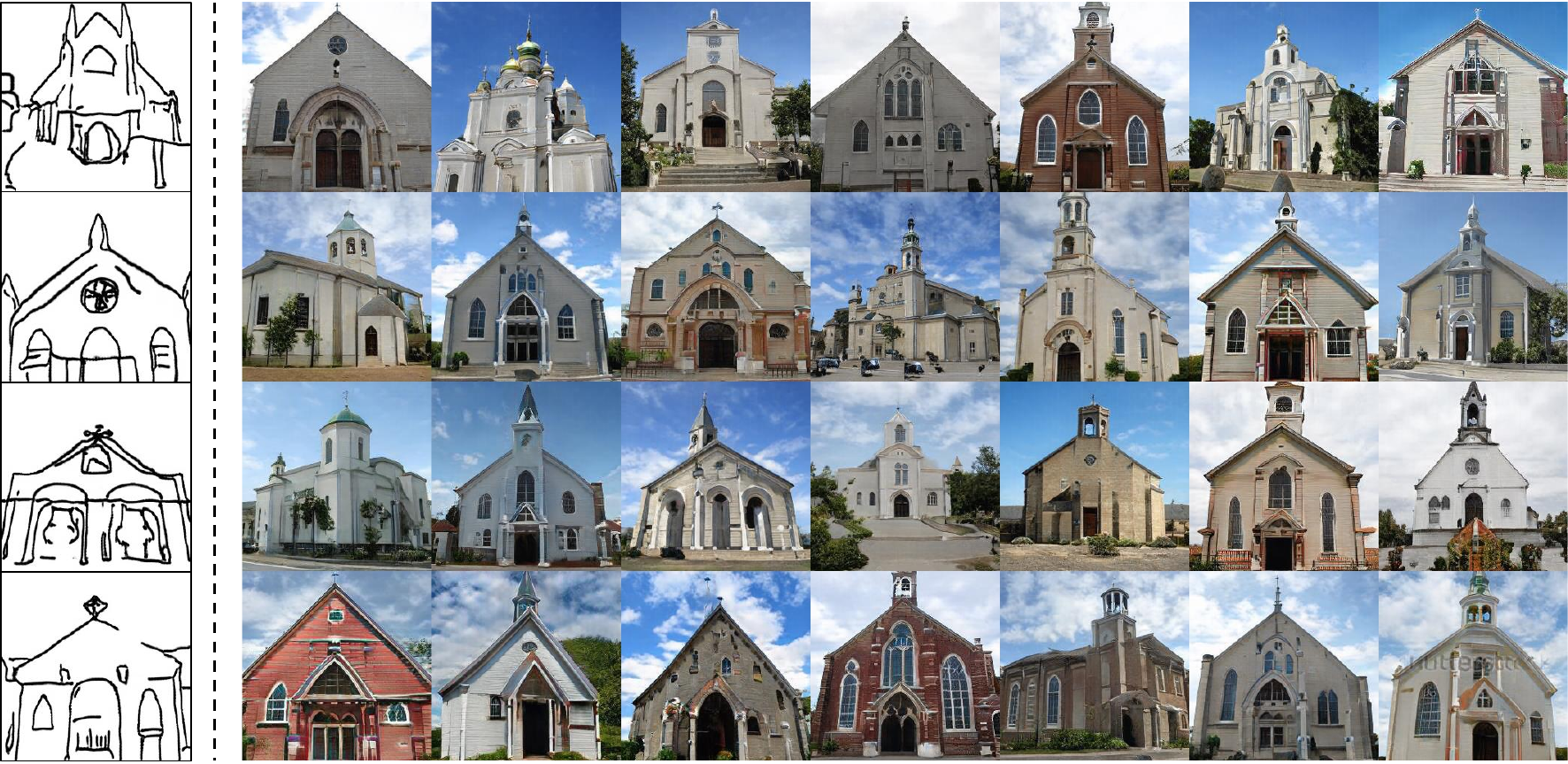}
    \caption{Uncurated samples of the \textbf{gabled church} model. Truncation $\psi=0.5$ is applied to generate the images.}    \label{fig:church_triangle}
\end{figure*}

\paragraph{Additional qualitative results.}
In \reffig{more_edits}, we show additional results on latent space editing with our customized models.
Also, we show uncurated samples for our models in \reffig{horse_ride} (horse rider), \reffig{horse_side} (horse on a side), \reffig{cat_stand} (standing cat) and \reffig{church_triangle} (gabled church). 

\section{Changelog}
\myparagraph{v1} Initial preprint release.

\myparagraph{v2} Updated customized FFHQ models and moved the corresponding figure to \reffig{ffhq}. Corrected the information in \reftbl{loss_weight}. Added experiments on interpolating two customized models (\reffig{model_interp}).

\begin{table}[t]
\centering
\begin{tabular}{lcc}
\toprule
 & $\lambda_{\text{image}}$ & $\lambda_{\text{weight}}$ \\ \midrule
$\losssketch$ & 0 & 0 \\
$\losssketch$+aug. & 0 & 0 \\
$\losssketch$+$\lossweight$ & 0 & 100 \\
$\losssketch$+$\lossweight$+aug. & 0 & 100 \\
\bottomrule
\end{tabular}
\vspace{1.5pt}
\caption{\textbf{Loss weights for each variant.} For a fair comparison, we use different loss weights for several variants. We find that using the above weights gives optimal performance. The variants not listed in this table is using the same hyperparameters as the \textbf{Full} method.}
\label{tbl:loss_weight}
\end{table}

\begin{table*}[t]
\centering
\resizebox{.9\linewidth}{!}{
\begin{tabular}{cccccccccccccccc}
\toprule
\multirow{4}{*}{Family} & \multirow{4}{*}{Name} & \multicolumn{2}{c}{Training settings} & \multicolumn{12}{c}{Test cases} \\ \cmidrule(lr){3-4} \cmidrule(lr){5-16} 
 &  & \multirow{3}{*}{\begin{tabular}[c]{@{}c@{}}No.\\ Samples\end{tabular}} & \multirow{3}{*}{Aug.} & \multicolumn{3}{c}{Horse rider} & \multicolumn{3}{c}{Horse on a side} & \multicolumn{3}{c}{Standing cat} & \multicolumn{3}{c}{Gabled church} \\ \cmidrule(lr){5-7} \cmidrule(lr){8-10}  \cmidrule(lr){11-13}  \cmidrule(lr){14-16}
 &  &  &  & PPL & Prec. & Rec. & PPL & Prec. & Rec. & PPL & Prec. & Rec. & PPL & Prec. & Rec. \\
 &  &  &  & $\downarrow$ & $\uparrow$ & $\uparrow$ & $\downarrow$ & $\uparrow$ & $\uparrow$ & $\downarrow$ & $\uparrow$ & $\uparrow$ & $\downarrow$ & $\uparrow$ & $\uparrow$ \\ \midrule
Pre-trained & Original & N/A &  & \gray{338.87} & \gray{0.22} & \gray{0.63} & \gray{338.87} & \gray{0.33} & \gray{0.57} & \gray{438.11} & \gray{0.21} & \gray{0.54} & \gray{342.73} & \gray{0.46} & \gray{0.49} \\ \midrule
\multirow{2}{*}{Baseline} & Bui~\etal~\cite{bui2017compact} & 30 &  & 356.56 & 0.24 & 0.53 & 343.48 & 0.26 & \textbf{0.60} & 433.05 & 0.22 & \textbf{0.58} & 346.48 & 0.49 & 0.48 \\
 & Chamfer & 30 &  & 353.07 & 0.30 & \textbf{0.56} & 371.11 & 0.35 & 0.57 & 418.91 & 0.26 & 0.55 & 340.12 & \textbf{0.50} & \textbf{0.52} \\ \midrule
\multirow{2}{*}{Ours} & Full (w/o aug.) & 30 &  & 353.71 & 0.42 & 0.52 & 266.69 & 0.42 & 0.49 & \textbf{150.89} & \textbf{0.65} & 0.20 & 344.24 & 0.48 & 0.48 \\
 & Full (w/ aug.) & 30 & \checkmark & \textbf{306.81} & \textbf{0.50} & 0.50 & \textbf{232.95} & \textbf{0.44} & 0.39 & 263.99 & 0.50 & 0.41 & \textbf{336.67} & 0.46 & 0.51 \\ \bottomrule
\end{tabular}}
\caption{{\bf Other metrics.} We report the Perceptual Path Length (PPL), Precision (Prec.), and Recall (Rec.) of the original models, baselines and our methods on four different test cases. The details of the baselines are in \refsec{quantitative_eval}. $\checkmark$ indicates translation augmentation is applied. $\uparrow$, $\downarrow$ indicate if higher or lower is better. Evaluations on the original models are in \gray{gray}, and the best value is highlighted in {\bf black}.}
\label{tbl:other_metrics}
\vspace{-5pt}
\end{table*}

\end{document}